\PassOptionsToPackage{nonamebreak}{natbib}
\PassOptionsToPackage{capitalize}{cleveref}
\documentclass{applemlr}

\newif\ifisarxiv
\isarxivtrue

\usepackage{amsmath,amsfonts,bm}

\def\eqref#1{equation~\ref{#1}}

\def\1{\bm{1}}

\newcommand{\test}{\mathcal{D_{\mathrm{test}}}}

\DeclareMathAlphabet{\mathsfit}{\encodingdefault}{\sfdefault}{m}{sl}
\SetMathAlphabet{\mathsfit}{bold}{\encodingdefault}{\sfdefault}{bx}{n}

\usepackage{amssymb}
\usepackage{subcaption}

\usepackage{hyperref}

\usepackage{url}
\usepackage{booktabs}
\usepackage{graphicx}
\usepackage{xcolor}
\usepackage[normalem]{ulem}
\usepackage{enumitem}
\usepackage{algorithm}
\usepackage{algpseudocode}
\usepackage{zref-savepos}
\usepackage{placeins}
\algrenewcommand\algorithmicrequire{\textbf{Input:}}
\algrenewcommand\algorithmicensure{\textbf{Output:}}
\algrenewcommand\Call[2]{\textproc{#1}\ifthenelse{\equal{#2}{}}{}{\left(#2\right)}}
\algrenewtext{Function}[2]{\algorithmicfunction\ \textproc{#1}\ifthenelse{\equal{#2}{}}{}{\ensuremath{\left(#2\right)}}}
\algrenewtext{If}[1]{\algorithmicif\hspace{0.4em}#1\hspace{0.4em}\algorithmicthen}
\algrenewtext{ElsIf}[1]{\algorithmicelse\ \algorithmicif\hspace{0.4em}#1\hspace{0.4em}\algorithmicthen}
\makeatletter
\renewcommand{\fnum@algorithm}{\fname@algorithm~\thealgorithm.}
\newcount\ALG@ruleid
\newcount\ALG@lineid
\newdimen\ALG@rulelength
\newif\ifALG@linepending
\newcommand{\ALG@finishline}{%
  \ifALG@linepending
    \unskip\zsaveposy{alg-line-end-\the\ALG@lineid}\par
    \ALG@linependingfalse
  \fi
}
\algrenewcommand\ALG@endalgorithmic{\ALG@finishline}
\newcommand{\PBPTalgend}[2]{%
  \expandafter\gdef\csname ALG@lastline@#1\endcsname{#2}%
}
\let\ALG@originalendblock\ALG@endblock
\renewcommand{\ALG@endblock}{%
  \ifdim\csname ALG@ind@\theALG@nested\endcsname>\z@
    \protected@write\@auxout{}{%
      \string\PBPTalgend{\csname ALG@rule@\theALG@nested\endcsname}{\the\ALG@lineid}}%
  \fi
  \ALG@originalendblock
}
\newcommand{\ALG@drawrule}[1]{%
  \zsaveposy{alg-block-start-#1}%
  \ifcsname ALG@lastline@#1\endcsname
    \edef\ALG@endlabel{alg-line-end-\csname ALG@lastline@#1\endcsname}%
    \zref@ifrefundefined{alg-block-start-#1}{}{%
      \zref@ifrefundefined{\ALG@endlabel}{}{%
        \ALG@rulelength=\dimexpr\zposy{alg-block-start-#1}sp-\zposy{\ALG@endlabel}sp\relax
        \smash{\rlap{\color[HTML]{C0C6CC}%
          \rule[\dimexpr-\ALG@rulelength-0.5\baselineskip+2pt\relax]{0.5pt}%
            {\dimexpr\ALG@rulelength+1.25\baselineskip-4pt\relax}}}%
      }%
    }%
  \fi
}
\newcommand{\ALG@putindents}{%
  \count@=\z@
  \loop\ifnum\count@<\theALG@nested
    \advance\count@\@ne
    \hskip0.3em
    \ifnum\csname ALG@first@\the\count@\endcsname=1
      \ALG@drawrule{\csname ALG@rule@\the\count@\endcsname}%
      \expandafter\let\csname ALG@first@\the\count@\endcsname\z@
    \fi
    \hskip\dimexpr\csname ALG@ind@\the\count@\endcsname-0.3em\relax
  \repeat
}
\def\ALG@doentity{%
  \ALG@finishline
  \edef\ALG@thisblock{\csname ALG@currentblock@\theALG@nested\endcsname}%
  \expandafter\ifx\csname ALG@b@\ALG@L @\ALG@thisentity @\ALG@thisblock\endcsname\relax
    \def\ALG@thisblock{0}%
  \fi
  \ALG@getentitytext
  \ifnum\ALG@thisblock=0\else\ALG@doend\fi
  \ifx\ALG@text\ALG@x@notext
    \item[]\nointerlineskip
  \else
    \item
    \global\advance\ALG@lineid\@ne
    \noindent\ALG@putindents
    \ALG@linependingtrue
  \fi
  \expandafter\ifnum0=\csname ALG@b@\ALG@L @\ALG@thisentity @\ALG@thisblock\endcsname\else
    \ALG@dobegin
    \global\advance\ALG@ruleid\@ne
    \expandafter\edef\csname ALG@rule@\theALG@nested\endcsname{\the\ALG@ruleid}%
    \expandafter\let\csname ALG@first@\theALG@nested\endcsname\@ne
  \fi
  \def\ALG@entitiecommand{\ALG@displayentity}%
}
\makeatother
\algtext*{EndIf}
\algtext*{EndWhile}
\algtext*{EndFor}
\algtext*{EndFunction}
\algblockdefx[ParFor]{ParFor}{EndParFor}[1]{\textbf{parallel for} #1 \textbf{do}}{}
\algtext*{EndParFor}
\newcommand{\act}[1]{\textit{#1}}

\title{Population-Based Post-Training for Large Language Models}

\author[1,*]{Yixuan Even Xu}
\author[2]{Aleksei Petrenko}
\author[2]{Raja Giryes}
\author[2]{Jackson Hamburger}
\author[2]{Kevin Chen}
\author[1]{Fei Fang}
\author[1]{\mbox{J. Zico Kolter}}
\author[2]{Philipp Kr\"ahenb\"uhl}
\affiliation[1]{Carnegie Mellon University}
\affiliation[2]{Apple}
\contribution[*]{Work performed during an internship at Apple.}
\correspondence{Yixuan Even Xu: \email{yixuanx@cs.cmu.edu}}

\abstract{
Reinforcement learning with verifiable rewards (RLVR) has brought us reasoning models, coding LLMs, and embodied vision-and-language agents.
RLVR typically trains a single model under hyperparameters chosen ahead of training, while researchers tune hyperparameters across consecutive runs.
We introduce \textbf{P}opulation-\textbf{B}ased \textbf{P}ost-\textbf{T}raining (PBPT), which trains several models in parallel and lets the training process do the tuning through population-based training.
PBPT periodically evaluates models on a validation set.
The process duplicates high performers and perturbs their hyperparameters to explore new configurations.
Low performers stop training in favor of high performers.
On reasoning benchmarks (Enigmata, Nemotron) and interactive tool-use environments (Gaia2, AppWorld), PBPT produces stronger agents than compute-matched groups of independent RLVR learners.
Furthermore, PBPT recovers from poor initial hyperparameters, automatically discovering better settings over the course of training.
}

\begin{document}
\maketitle

\section{Introduction}
\label{sec:intro}

Reinforcement learning with verifiable rewards (RLVR) elicits strong reasoning behavior from a pretrained large language model (LLM) without supervised reasoning traces \citep{DBLP:journals/corr/abs-2501-12948}.
However, RLVR training can be unstable: a collapsed learner may not recover through continued training alone \citep{DBLP:journals/corr/abs-2507-18071}.
Moreover, standard RLVR fixes training configurations before training begins, even though effective hyperparameters may change over time. 
Population-based training (PBT)~\citep{DBLP:journals/corr/abs-1711-09846} addresses both problems: it trains multiple agents concurrently, periodically copies the best-performing agents' weights to weaker peers, and perturbs their hyperparameters.
As a result, PBT restarts stagnating learners from peers that are still improving and adapts configurations over training.
Despite its success in deep reinforcement learning, PBT has seen little use in LLM post-training~\citep{DBLP:journals/corr/abs-2605-16727}.

\begin{figure}[tbp]
  \centering
  \begin{tabular}{@{}c@{\hspace{0.029\textwidth}}c@{\hspace{0.029\textwidth}}c@{}}
    {\includegraphics[width=0.450\textwidth]{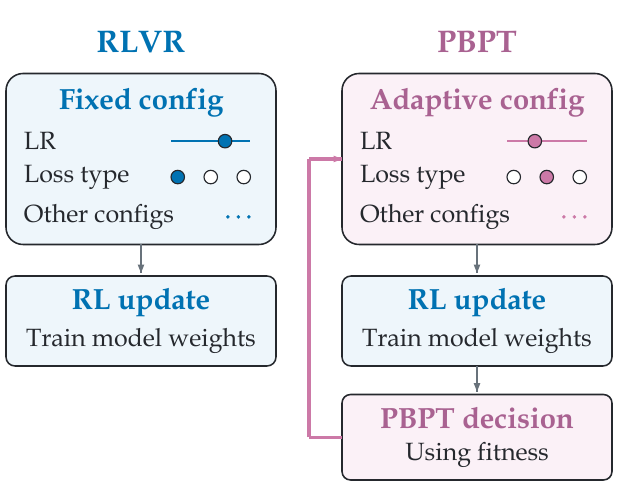}}
    & {\color[HTML]{D9DEE2}\rule{0.35pt}{0.3533\textwidth}}
    & \includegraphics[width=0.4893\textwidth]{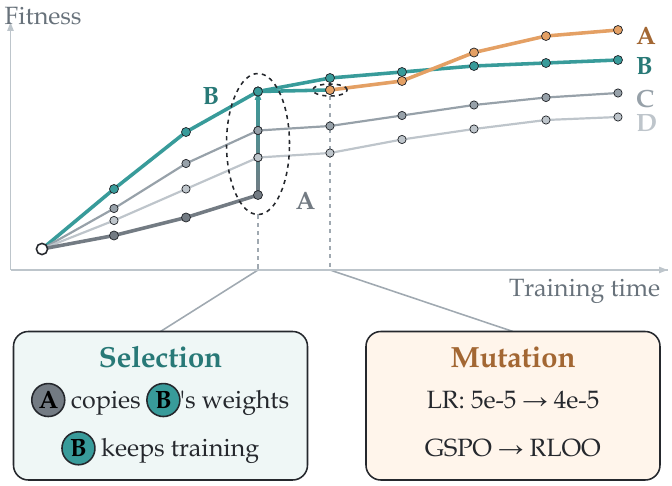} \\[5pt]
    {\bfseries (a) Population-based post-training}
    && {\bfseries (b) PBPT decisions in a population}
  \end{tabular}
  \caption{PBPT enables parallel RLVR learners to share progress while adapting their training configurations online.
  Unlike independent RLVR with a fixed configuration, PBPT adds an outer optimization loop guided by validation fitness, as shown in Subfigure~(a).
  Two mechanisms, \textit{selection} and \textit{mutation}, coordinate this joint optimization, as illustrated separately in Subfigure~(b).
  Selection copies weights from the higher-fitness learner B to A.
  Mutation changes A's configuration while preserving its weights, allowing A to overtake B through subsequent training.}
  \label{fig:teaser}
\end{figure}

This work provides one of the first systematic studies of PBT for a population of LLM learners trained with RLVR on the same task.
We introduce \textbf{PBPT} (\textbf{P}opulation-\textbf{B}ased \textbf{P}ost-\textbf{T}raining), a decentralized framework that adds an outer loop for online hyperparameter optimization to parallel RLVR training.
See \cref{fig:teaser} for an overview.
Concretely, PBPT uses a shared filesystem as a lightweight communication medium that exposes the state of the population to all learners.
Individual learners train asynchronously and periodically query this shared state.
When underperforming, a learner may reinitialize from a stronger peer through a \textit{selection} mechanism.
An additional \textit{mutation} mechanism perturbs training configurations, allowing the population to continue exploring the hyperparameter space while successful configurations propagate through selection.

We evaluate PBPT on reasoning and multi-turn agentic tool-use benchmarks: Enigmata \citep{DBLP:conf/nips/ChenHYCCDYCLYZW25}, Nemotron \citep{DBLP:journals/corr/abs-2512-20848}, Gaia2 \citep{DBLP:journals/corr/abs-2602-11964}, and AppWorld \citep{DBLP:conf/acl/TrivediKHMDLGSB24}.
Across all four, PBPT outperforms independent RLVR with the same base model, number of learners, and total compute budget.
Further experiments on Enigmata show that PBPT recovers and keeps improving from suboptimal initial configurations where independent RLVR degrades substantially.


\section{Related Work}
\label{sec:related_work}

\textbf{Reinforcement learning with verifiable rewards} (RLVR) post-trains LLMs using automatically verifiable rewards and elicits reasoning without supervised solution traces \citep{DBLP:journals/corr/abs-2501-12948}.
GRPO is a widely used foundation for reasoning-oriented RLVR, estimating advantages from groups of sampled responses without a learned critic \citep{DBLP:journals/corr/abs-2402-03300}.
Related objectives include REINFORCE-style estimators such as RLOO \citep{DBLP:conf/acl/AhmadianCGFKPUH24} and sequence-level clipped objectives such as GSPO \citep{DBLP:journals/corr/abs-2507-18071}.
Subsequent variants target training stability, credit assignment, and data or compute efficiency, such as DAPO, VinePPO, VAPO, and PODS \citep{DBLP:conf/nips/YuZZYZYDFLLLLLL25,DBLP:conf/icml/KazemnejadAPSRC25,DBLP:journals/corr/abs-2504-05118,DBLP:journals/tmlr/XuSFK26}.
PBPT complements these methods: rather than introducing a new optimization objective, it orchestrates a population of independent RLVR learners that continuously adapt their hyperparameter configurations.
In this work, PBPT uses GSPO and RLOO as its underlying RL objectives, although it supports any set of RLVR objectives and can switch between them during training.

\textbf{Population-based training} (PBT) jointly optimizes model parameters and hyperparameter schedules by replacing weak learners with perturbed copies of stronger ones \citep{DBLP:journals/corr/abs-1711-09846}.
In game-playing tasks, PBT has achieved human-level performance in Quake III Capture the Flag and grandmaster-level play in StarCraft II \citep{doi:10.1126/science.aau6249,DBLP:journals/nature/VinyalsBCMDCCPE19}.
Subsequent work generalizes PBT through flexible asynchronous frameworks and Bayesian search over mixed hyperparameter spaces and network architectures \citep{DBLP:conf/kdd/LiSPDJGBHG19,DBLP:conf/nips/Parker-HolderND21,DBLP:conf/automl/WanLPBNRO22}.
Recent population-oriented LLM methods train interacting policies through cooperative or collective reinforcement learning \citep{DBLP:conf/nips/MaHPLA0C24,DBLP:journals/corr/abs-2508-12338,DBLP:journals/corr/abs-2510-11062} or evolve model weights without gradient updates \citep{DBLP:conf/acl/ZhangYYFZBOH26}.
These settings differ from ours because they either assign manually designed roles to population members or do not employ RLVR updates.
In LLM post-training literature, PopuLoRA \citep{DBLP:journals/corr/abs-2605-16727} is closest to our setting.
It co-evolves asymmetric teacher and student LoRA populations for reasoning self-play rather than training homogeneous RLVR learners.
Algorithmically, PBPT is closest to DexPBT, which also lets independent RL learners make selection and mutation decisions without a central coordinator \citep{DBLP:conf/rss/PetrenkoASHM23}.
PBPT adapts this decentralized framework to the LLM post-training setting.

\textbf{Prompt and harness optimization.}
Prompt evolution methods use evaluation or trajectory feedback to search over textual candidates while keeping model weights fixed \citep{DBLP:conf/iclr/Yang0LLLZC24,DBLP:conf/iclr/Guo0GLS0L0Y24,DBLP:conf/icml/FernandoBMOR24,DBLP:journals/corr/abs-2507-19457,DBLP:journals/corr/abs-2606-04465,DBLP:conf/cais/AgrawalLTMESSKS26}.
Related methods search over executable agent designs and harnesses, including ADAS, DGM, and TTHE \citep{DBLP:conf/iclr/HuLC25,DBLP:journals/corr/abs-2505-22954,DBLP:journals/corr/abs-2607-08124}.
Recent joint optimization methods such as E-SPL, INSPO, $P^2O$, and WHALE are able to optimize prompts or executable harnesses alongside an updating LLM policy \citep{DBLP:journals/corr/abs-2602-14697,DBLP:journals/corr/abs-2512-01945,DBLP:journals/corr/abs-2603-21877,kim2026whalesimplerecipejoint}.
Mutation in PBPT may similarly include non-parametric harness optimization, placing it within this broader literature, although such optimization is not the focus of this work.

\textbf{Agentic control.}
Recent work, including Autofinetune, PostTrainBench, AutoTrainess, and EvoTrainer, studies LLM agents that use experimental feedback to guide LLM post-training \citep{wei2026autofinetune,DBLP:journals/corr/abs-2603-08640,DBLP:journals/corr/abs-2606-31551,DBLP:journals/corr/abs-2606-03108}.
PBPT also adapts training configurations based on evaluation feedback, but coordinates parallel learners through selection and hyperparameter mutation rather than asking an LLM agent to decide on the training changes.

\section{Preliminaries}
\label{sec:prelim}

\textbf{Autoregressive language models } factorize a distribution $p_\theta\left(\mathbf{x}\mid\mathbf{c}\right)=\prod_{t=1}^{T}p_\theta\left(x_t\mid\mathbf{c},x_{1:t-1}\right)$ over a generated sequence $\mathbf{x}=\left(x_1,\ldots,x_T\right)\in\mathcal{V}^{T}$ given its context $\mathbf{c}\in\mathcal{V}^*$.
Here, $\theta$ denotes the parameters of the language model, and $\mathcal{V}$ is its vocabulary.

\textbf{Reinforcement learning with verifiable rewards} (RLVR)
samples a response or tool-use trajectory $\mathbf{x}\sim p_\theta\left(\cdot\mid\mathbf{c}\right)$ from a context $\mathbf{c}$ of a task distribution $\mathcal{T}$.
A verifier then automatically assigns each sample a terminal reward $R\left(\mathbf{c},\mathbf{x}\right)$.
The objective is to maximize $J\left(\theta\right)=\mathbb{E}_{\mathbf{c}\sim\mathcal{T},\mathbf{x}\sim p_\theta\left(\cdot\mid\mathbf{c}\right)}[R\left(\mathbf{c},\mathbf{x}\right)]$.
Policy gradient methods such as REINFORCE leave-one-out (RLOO) \citep{DBLP:conf/acl/AhmadianCGFKPUH24}, group-relative policy optimization (GRPO)~\citep{DBLP:journals/corr/abs-2402-03300}, and group sequence policy optimization (GSPO) \citep{DBLP:journals/corr/abs-2507-18071} optimize this objective using sample-based advantage estimates.
RLVR relies on a well-chosen set of hyperparameters $h$: the RL objective, learning rate, trust-region clipping, rollout budget, etc.
Poor configurations can cause training to stall or collapse.

\paragraph{Population-based training.}
PBT optimizes hyperparameters and stabilizes training by maintaining a population $\mathcal{P}=\{\left(\theta_i,h_i\right)\}_{i=1}^{n}$ of concurrent learners, where $\theta_i$ denotes model parameters and $h_i$ denotes training hyperparameters \citep{DBLP:journals/corr/abs-1711-09846}.
Each member alternates between several optimization steps under its current hyperparameters and population updates based on available evaluation scores.
Three terms are useful for describing this process: \textit{fitness}, \textit{selection}, and \textit{mutation}.

\textit{Fitness}
$\mathrm{fitness}_i=\Call{Eval}{\theta_i}$ of member $i$ is a scalar-valued evaluation score for its current checkpoint.
Higher values indicate a better checkpoint.
Unlike the RLVR reward $R\left(\mathbf{c},\mathbf{x}\right)$ for an individual trajectory, $\mathrm{fitness}_i$ compares checkpoints during population updates.
Standard PBT compares members at their latest wall-clock checkpoints \citep{DBLP:journals/corr/abs-1711-09846}, while DexPBT compares historical checkpoints at equal environment steps \citep{DBLP:conf/rss/PetrenkoASHM23}.
In LLM post-training, hyperparameter changes alter per-step compute and iteration times by orders of magnitude.
PBPT instead compares peers at matched accumulated compute rather than equal steps or wall-clock time.

\textit{Selection}
replaces a lower-fitness member $i$ (the recipient) with a higher-fitness member $j$ (the donor).
The recipient copies the donor's model parameters, optimizer state, and hyperparameters.
Standard PBT changes hyperparameters only when it replaces weights \citep{DBLP:journals/corr/abs-1711-09846}.
DexPBT mutates middle-ranked learners in-place and replaces the bottom fraction by rank alone \citep{DBLP:conf/rss/PetrenkoASHM23}.
Replacing LLM weights too often reduces population diversity.
PBPT instead uses fitness-gap thresholds to choose between in-place mutation and donor replacement.

\textit{Mutation}
changes a member's hyperparameters during training through a mutation operator $\Call{Mutate}{h_i}$.
Standard PBT multiplies a hyperparameter by $0.8$ or $1.2$ or resamples it from a prior \citep{DBLP:journals/corr/abs-1711-09846}, while DexPBT scales hyperparameters by a random factor \citep{DBLP:conf/rss/PetrenkoASHM23}.
PBPT applies crossover and mutation to both continuous and discrete RLVR settings, including the RL objective, rollout budgets, trust-region clipping, and task sampling.

\section{Population-Based Post-Training}
\label{sec:method}


\begin{algorithm}[htbp]
\caption{Population-Based Post-Training (PBPT)}
\label{alg:pbpt}
\begin{algorithmic}[1]
\Require Initial population $\left\{\mathrm{state}_i\right\}_{i=1}^n$, training and validation data, per-member budget
\Ensure Trained population $\left\{\mathrm{state}_i\right\}_{i=1}^n$
\State $\mathcal{S}\gets\varnothing$ \Comment{Shared record set (state, fitness, time)}
\ParFor{member $i\in\left\{1,\ldots,n\right\}$} \Comment{Asynchronously}
    \For{evaluation $k=1,2,\ldots$} \Comment{Until training budget is reached}
        \State $\mathrm{state}_i\gets\Call{Train}{\mathrm{state}_i}$ \Comment{RLVR until evaluation $k$}
        \State $\mathrm{fitness}_{i,k}\gets\Call{Eval}{\mathrm{state}_i}$ \Comment{Validation fitness}
        \State $\mathcal{S}\gets\mathcal{S}\cup\bigl\{\underbrace{\left(\mathrm{state}_i,\mathrm{fitness}_{i,k},\mathrm{time}_{i,k}\right)}_{\mathcal{S}_{i,k}}\bigr\}$ \Comment{With cumulative job runtime $\mathrm{time}_{i,k}$}
        \State $\mathcal{C}_{i,k}\gets\left\{\operatorname{most\ recent}_{\mathrm{time}_{j,\hat{k}}\le\mathrm{time}_{i,k}}\mathcal{S}_{j,\hat{k}}\;\middle|\;j\in\left\{1,\ldots,n\right\}\right\}$ \Comment{Compute-matched peers}
        \State $\left(\mathrm{action},\mathrm{donor}\right)\gets\Call{Decide}{i,\mathcal{C}_{i,k},\mathcal{S}}$
        \If{$\mathrm{action}=\act{replace}$} \Comment{Copy donor weights, crossover config}
            \State $\left(\theta_i,\omega_i\right)\gets\left(\theta_{\mathrm{donor}},\omega_{\mathrm{donor}}\right)$ and $\Call{Crossover}{h_i,h_{\mathrm{donor}}}$
        \EndIf
        \If{$\mathrm{action}\in\left\{\act{replace},\act{perturb}\right\}$} \Comment{Mutate configuration}
            \State $\Call{Mutate}{h_i}$
        \EndIf
    \EndFor
\EndParFor
\State \Return $\left\{\mathrm{state}_i\right\}_{i=1}^n$
\end{algorithmic}
\end{algorithm}

\label{subsec:pbpt_overview}

\textbf{P}opulation-\textbf{B}ased \textbf{P}ost-\textbf{T}raining (\cref{alg:pbpt}) trains $n$ LLM policies in parallel.
Each member $i$ has a training state $\mathrm{state}_i=\left(\theta_i,\omega_i,h_i\right)$ that contains its policy parameters $\theta_i$, optimizer state $\omega_i$, and hyperparameters $h_i$ (such as the learning rate, rollouts per task, and RL objective).
Between evaluations, member $i$ collects rollouts and updates its policy with RLVR $(\Call{Train}{\mathrm{state}_i})$.
At evaluation $k$, member $i$ evaluates its checkpoint on the validation split $(\Call{Eval}{\mathrm{state}_i})$, writes the record $\mathcal{S}_{i,k}=\left(\mathrm{state}_i,\mathrm{fitness}_{i,k},\mathrm{time}_{i,k}\right)$ at cumulative job runtime $\mathrm{time}_{i,k}$ to shared storage $\mathcal{S}$, and builds a peer comparison set $\mathcal{C}_{i,k}\subseteq\mathcal{S}$.
Using $\mathcal{C}_{i,k}$, \textsc{Decide} chooses one of three actions (\cref{alg:pbpt_operators} in \cref{app:implementation_details}): \act{keep} leaves the state unchanged, \act{perturb} mutates $h_i$ in-place, and \act{replace} copies a donor's weights and optimizer state, crosses over $h_i$ with $h_{\mathrm{donor}}$, then mutates $h_i$.

PBPT faces three main challenges.
First, PBPT trains all its LLM policies in parallel, and completely asynchronously.
Second, RLVR, and thus PBPT, operates on noisy evaluations of the fitness function.
Third, RLVR presents a complex search space with many interacting hyperparameters, including both continuous and discrete choices.

\paragraph{Asynchronous execution and compute-matched peer comparisons.}
\label{subsec:async_execution}
Different policies in PBPT train at different rates, depending on cluster configurations, hyperparameters, or simply noise in the training and rollout generation.
For example, policies with smaller batch sizes train faster than policies with larger batch sizes or bigger rollout buffers.
Rollouts with higher temperatures may lead to more diverse batches, but can also lead to longer rollouts and thus slower rollout collection and gradient updates.
Comparing the latest checkpoints, as in standard asynchronous PBT, favors members that use more compute rather than those that use compute more efficiently.
Synchronizing iterations across hyperparameter settings is nearly impossible.
PBPT instead designs its \textit{fitness} and \textit{selection} criteria to be compute-dependent.
Specifically, each member makes population decisions on its own using records in shared storage $\mathcal{S}$, without a central coordinator.
PBPT matches peers with a clock $\mathrm{time}_{i,k}$ that measures active job runtime (including training and evaluation while active, and pausing while stopped).
At evaluation $k$, member $i$ builds $\mathcal{C}_{i,k}$ from each peer $j$'s most recent record $\mathcal{S}_{j,\hat{k}}\in\mathcal{S}$ with $\mathrm{time}_{j,\hat{k}}\le\mathrm{time}_{i,k}$.
To ignore stalled jobs, member $i$ excludes any peer whose latest eligible record is more than $2$ hours behind in both runtime and wall-clock time (\cref{app:implementation_details}).
When a member copies a donor checkpoint during \act{replace}, it keeps its own clock $\mathrm{time}_i$.
This compute-matched peer comparison allows PBPT to fairly compare, select, and donate weights across multiple asynchronously training policies.

\paragraph{Selection and state transfer.}
\label{subsec:selection}
Validation scores in LLM RLVR are noisy.
To keep the computation between training and validation balanced, validation runs are often quite small, and randomness in sampling of rollouts easily influences their verifiable reward.
PBPT thus uses a much stricter \textit{replacement} strategy compared to regular population-based training.
We only mutate the bottom-25\% of policies at any given time, compared to their compute-matched peers.
If a member's $\mathrm{fitness}_{i,k}$ lags behind its best-performing peer by more than $\tau_{\mathrm{mut}}$, its hyperparameters are \textit{perturbed}.
If a member's $\mathrm{fitness}_{i,k}$ lags behind its best-performing peer by more than $\tau_{\mathrm{sel}} > \tau_{\mathrm{mut}}$, it is \textit{replaced}.
%
The \act{replacement} operation chooses uniformly at random from a donor pool $\mathcal{D}$ of current best-performing peers (top $30\%$) and up to $10$ historical records matching or outperforming the best peer.
\textsc{Crossover} merges the donor's hyperparameters with the current member's, sampling each with probability $1/2$.
Any \act{replaced} member is also mutated.
PBPT selection ensures only members that lag significantly behind are \act{replaced}.
Low-performing members close to top-performers maintain their chances to improve with \act{perturbed} hyperparameters.

\paragraph{Configuration mutation.}
\label{subsec:mutation}
PBPT represents a configuration of $d$ mutable hyperparameters as $h=\left(h_1,\ldots,h_d\right)\in\mathcal{H}=\mathcal{H}_1\times\cdots\times\mathcal{H}_d$, where each hyperparameter $h_q$ has a declared type and valid range $\mathcal{H}_q$.
Hyperparameters include optimization algorithms, rollout collection, trust-region clipping, task sampling, and scenario replay (\cref{app:initial_configs}).
\textsc{Mutate} changes each hyperparameter $h_q\in h$ independently with probability $p_{\mathrm{mut}}$.
For numeric hyperparameters, \textsc{Mutate} samples $\mu\sim\operatorname{Uniform}[\mu_{\min},\mu_{\max}]$ and $s\sim\operatorname{Uniform}(\{-1,+1\})$, scales $h_q$ by $\mu^s$, and clips the result to $\mathcal{H}_q$.
For Boolean and categorical hyperparameters, \textsc{Mutate} samples uniformly from $\mathcal{H}_q\setminus\{h_q\}$ (\cref{app:implementation_details}).
Before training starts, PBPT also applies \textsc{Mutate} to a subset of members to create different initial configurations.

\FloatBarrier

\section{Experiments}
\label{sec:experiments}

In this section, we conduct experiments to evaluate the effectiveness of PBPT in LLM post-training.
We focus on reasoning and agentic tool-use tasks, using compute-matched independent RLVR runs as the main baseline.
We also show that PBPT adapts from suboptimal initial RL configurations and use ablations to establish that mutation is needed for PBPT to achieve its full gains.

\paragraph{Benchmarks.}
We consider four benchmarks spanning reasoning and agentic tool use: Enigmata, Nemotron, Gaia2, and AppWorld.

\emph{Enigmata} \citep{DBLP:conf/nips/ChenHYCCDYCLYZW25} consists of $36$ puzzle families covering constraint satisfaction, search, symbolic reasoning, and logic.
We train on a cleaned version of the benchmark, correcting invalid problems and filtering out selected malformed task categories; see \cref{app:benchmark_details} for details.
We generate training instances using the official script, hold out a shuffled subset for validation, and sample the test set from Enigmata-Eval.
After this cleanup, the resulting splits contain $161{,}213$ training, $229$ validation, and $998$ test instances across $35$ families.

\emph{Nemotron} \citep{DBLP:journals/corr/abs-2512-20848} refers to NVIDIA's Nemotron-3-Nano RL training blend.
We use $6$ task subsets covering STEM multiple-choice questions, competition mathematics, instruction following, structured JSON generation, and competitive programming.
We split the data into disjoint sets, using $81{,}725$ tasks for training and $512$ tasks each for validation and test.

\emph{Gaia2} \citep{DBLP:journals/corr/abs-2602-11964} evaluates multi-turn agents on personal-assistant tasks involving applications such as email, calendar, contacts, and files.
We use its $800$ public scenarios across five capability categories: execution, search, adaptability, time, and ambiguity.
We split the data by \emph{universe}, an application environment with its own user data, to keep the underlying application data disjoint across splits.
This gives $458$ training, $153$ validation, and $189$ test scenarios.

\emph{AppWorld} \citep{DBLP:conf/acl/TrivediKHMDLGSB24} evaluates interactive coding agents that complete everyday tasks by writing and executing code against application APIs.
We use the $90$ training tasks and $55$ of the $57$ development tasks for validation, excluding $2$ development tasks with bugs in their definitions.
We evaluate on $168$ normal and $417$ challenge test tasks.

\begin{figure}[!t]
  \centering
  \includegraphics[width=\textwidth]{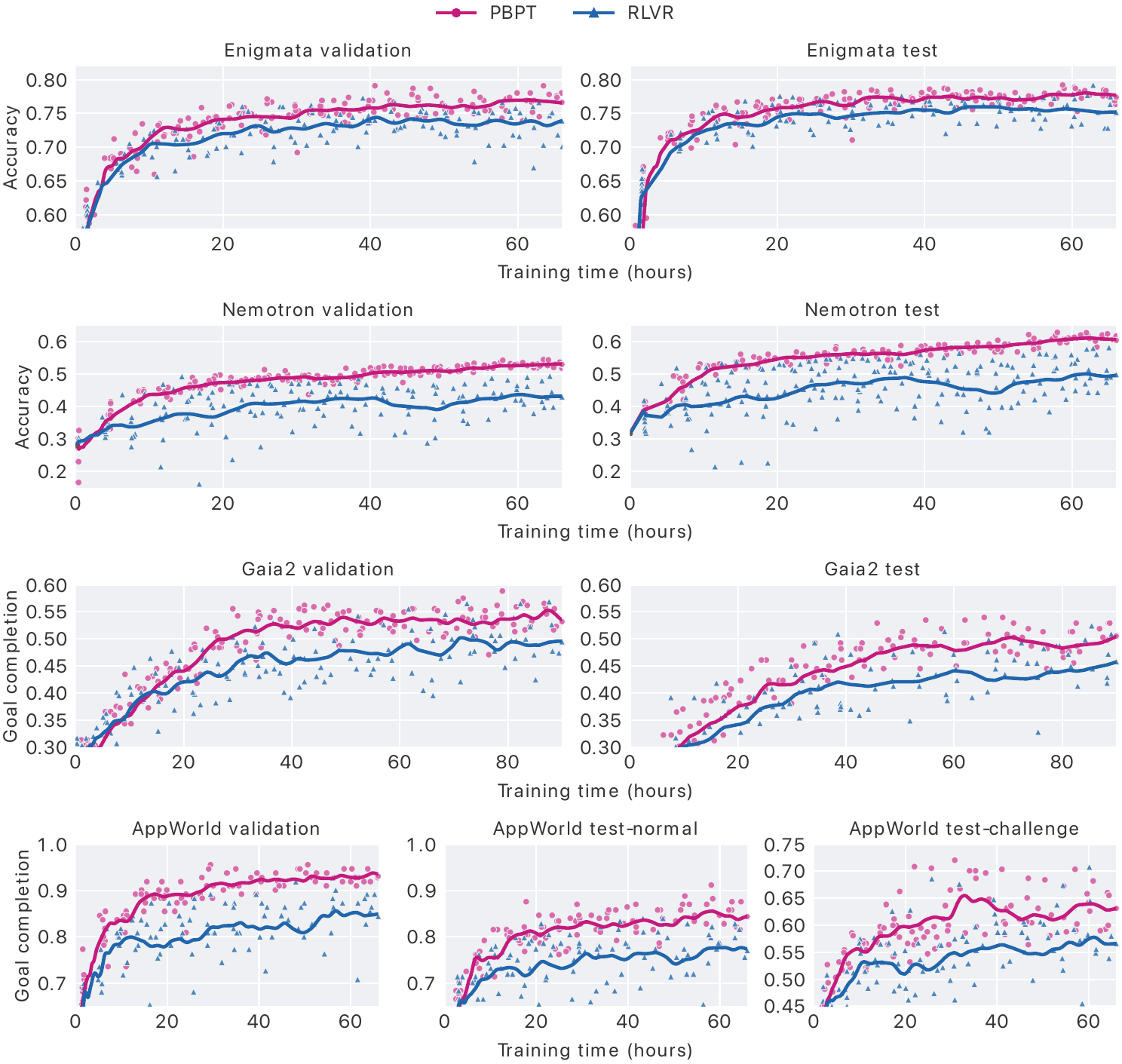}
  \caption{Comparison of PBPT and independent RLVR on Enigmata, Nemotron, Gaia2, and AppWorld.
  The x-axis shows accumulated training time per learner.
  The y-axis shows accuracy on Enigmata and Nemotron, and task goal completion on Gaia2 and AppWorld.
  Scatter points show subsampled evaluation results, and solid lines show smoothed means of the latest available evaluation from each population member or independent seed.
  The shared point at zero training time reports a separate evaluation of the base model.
  Compared with independent RLVR, PBPT achieves higher mean validation and test performance under matched training budgets.}
  \label{fig:main_results}
\end{figure}

\paragraph{Algorithms.}
We compare a PBPT population of $8$ learners with $8$ independent RLVR runs.
Both methods start from the same base model and use the same per-learner training resources.
For the main experiments, we manually tuned each benchmark's initial RL configuration over successive trial training runs.
Learners use different random seeds and share an RL configuration before PBPT's initial mutation.
PBPT may also change this configuration during training.
\Cref{app:benchmark_details} describes the training rewards, and \cref{app:initial_configs} gives the initial RL configurations and PBPT search spaces.

\paragraph{Evaluation.}
We use the training split for RLVR updates, the validation split for population decisions, and the test split only for reporting.
We report accuracy on Enigmata and Nemotron and task goal completion on Gaia2 and AppWorld, defined as the fraction of tasks completed successfully.
Success requires all task checks to pass.
\Cref{app:benchmark_details} provides data and evaluation details.

\paragraph{Implementation.}
We use Qwen3.5-4B for Enigmata, Nemotron, and AppWorld, and Qwen3.5-9B for Gaia2.
We use low-rank adaptation (LoRA) \citep{DBLP:conf/iclr/HuSWALWWC22} with rank $8$ and scaling factor $16$.
We use Accelerate \citep{accelerate} for distributed training and vLLM \citep{DBLP:conf/sosp/KwonLZ0ZY0ZS23} for rollout generation.
Each learner uses $4$ NVIDIA B200 GPUs for training, rollout generation, and evaluation.
We compare the methods over the first $66$ hours of training per learner on Enigmata, Nemotron, and AppWorld, and $90$ hours on Gaia2.
We measure training time as each learner's accumulated training-iteration time, including rollout collection and policy updates but excluding startup and intervals when training is stopped (see \Cref{app:implementation_details} for additional details).

\subsection{Performance Against Compute-Matched Independent RLVR}
\label{subsec:exp_main}

We present the training curves of PBPT and independent RLVR in \cref{fig:main_results}.
Across all four benchmarks, PBPT achieves higher mean validation and test performance than independent RLVR under matched training budgets.
PBPT maintains its advantage through the later stages of training.

The gains extend from validation to the held-out test splits, which do not inform population decisions.
Therefore, coordinating parallel learners with PBPT can improve RLVR post-training across reasoning and multi-turn tool-use tasks under matched training budgets.
\Cref{app:nemotron_9b} reports an additional comparison on Nemotron with Qwen3.5-9B.

\subsection{Adaptation from Suboptimal RL Configurations}
\label{subsec:res_adaptation}

RL hyperparameters can be difficult to tune.
Seemingly reasonable choices may lead to overly aggressive updates or unstable training \citep{DBLP:conf/nips/YuZZYZYDFLLLLLL25,DBLP:journals/corr/abs-2507-18071}.
We therefore evaluate whether PBPT can adapt from a plausible but suboptimal initial configuration.
On Enigmata, we manually choose values within PBPT's search ranges, including a higher learning rate, wider clipping bounds, more optimization epochs per iteration, and a smaller minibatch than in the well-tuned initial configuration (\cref{app:suboptimal_config}).
We compare $8$ independent RLVR runs with $8$ PBPT learners, all starting from this configuration with the same base model and per-learner resources.

\begin{figure}[tbp]
  \centering
  \includegraphics[width=\textwidth]{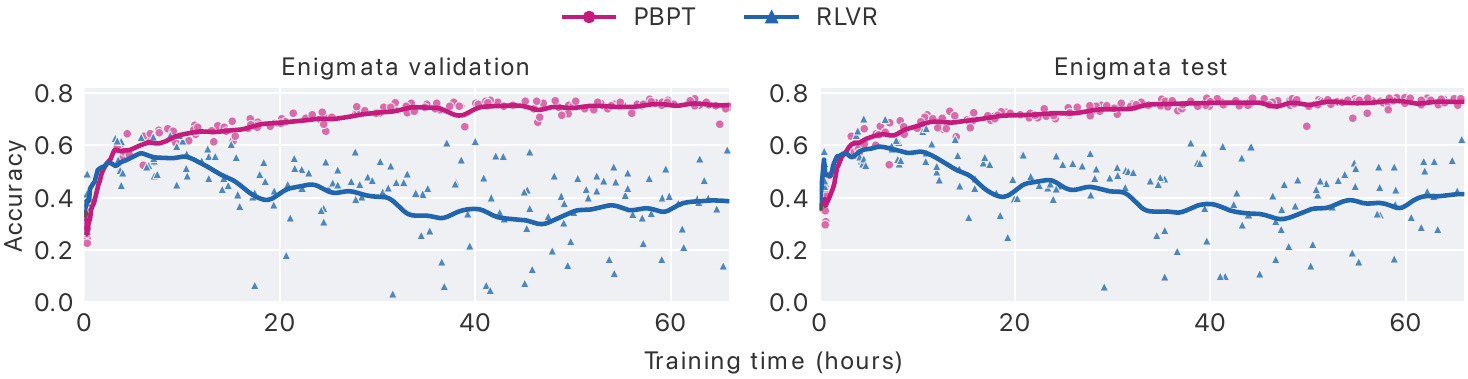}
  \caption{Comparison of PBPT and independent RLVR on Enigmata with a suboptimal initial configuration.
  The axes, points, and lines follow the same convention as in \cref{fig:main_results}.
  Over the course of training, PBPT continues to improve while independent RLVR loses its early gains.}
  \label{fig:adaptation_results}
\end{figure}

\Cref{fig:adaptation_results} shows that independent RLVR initially improves but then degrades substantially as training continues and never recovers.
In contrast, PBPT continues to improve, maintaining its gains throughout training.
These results show that PBPT can sustain policy improvement by searching for effective configurations during training, even when starting from a suboptimal initial configuration.

\subsection{The Role of Mutation}
\label{subsec:res_ablation}

PBPT combines selection with hyperparameter mutation.
To isolate the contribution of mutation, we evaluate a selection-only variant, \emph{PBPT without mutation}, which disables mutation both at initialization and throughout training.

We compare full PBPT, PBPT without mutation, and independent RLVR on Enigmata under both the tuned initial configuration from \cref{subsec:exp_main} and the suboptimal initial configuration from \cref{subsec:res_adaptation}.
Each method uses $8$ learners with the same base model and per-learner resources.

\begin{figure}[tbp]
  \centering
  \includegraphics[width=\textwidth]{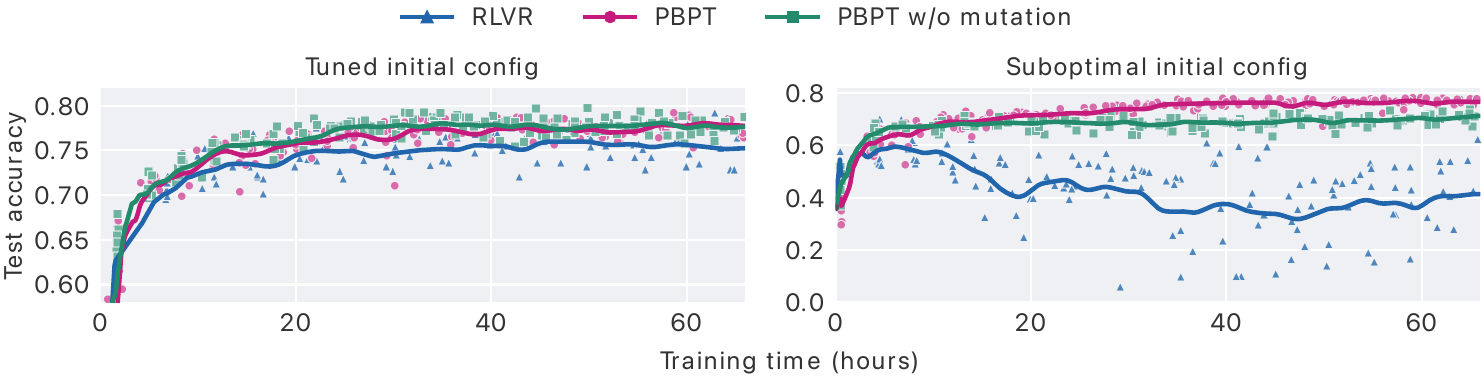}
  \caption{Effect of mutation on Enigmata test performance under tuned and suboptimal initial configurations.
  The axes, points, and lines follow the same convention as in \cref{fig:main_results}.
  PBPT without mutation underperforms full PBPT under the suboptimal configuration, but performs comparably under the tuned configuration.
  Both PBPT variants outperform independent RLVR.
  }
  \label{fig:ablation_results}
\end{figure}

As shown in \cref{fig:ablation_results}, the effect of mutation depends on the initial configuration.
With the tuned configuration, selection-only PBPT performs comparably to full PBPT, and both outperform independent RLVR.
With the suboptimal configuration, selection alone also prevents the large degradation observed in independent RLVR, but reaches a lower test accuracy than full PBPT.
We therefore see that mutation is necessary for PBPT's full gains from suboptimal initial configurations.

\subsection{Population-Level and Individual Performance}
\label{subsec:res_population}

We summarize in \cref{tab:enigmata_summary} the Enigmata runs with both tuned and suboptimal initial configurations over the final $20\%$ of the training budget.
We first average each member's validation and test evaluations within this window.
We then report the mean test accuracy across all $8$ members and the test means of the members with the highest and lowest validation means.

\begin{table}[t]
  \centering
  \setlength{\belowcaptionskip}{4pt}
  \caption{Enigmata test accuracy ($\%$) under tuned and suboptimal initial RL configurations, averaged over the evaluation results in the final $20\%$ of the training budget.
  Avg. test weights all $8$ members equally.
  Best-val. test and Worst-val. test report the test means of the members with the highest and lowest validation means, respectively.
  Bold marks the highest value in each column.}
  \label{tab:enigmata_summary}
  \setlength{\tabcolsep}{3.5pt}
  \renewcommand{\arraystretch}{1.12}
  \begin{tabular*}{\textwidth}{@{\extracolsep{\fill}}lccc@{\hspace{14pt}}ccc@{}}
    \toprule
    & \multicolumn{3}{c}{Tuned initial configuration} & \multicolumn{3}{c}{Suboptimal initial configuration} \\
    \cmidrule(lr){2-4} \cmidrule(l){5-7}
    Method & \shortstack{Avg.\\test} & \shortstack{Best-val.\\test} & \shortstack{Worst-val.\\test}
    & \shortstack{Avg.\\test} & \shortstack{Best-val.\\test} & \shortstack{Worst-val.\\test} \\
    \midrule
    RLVR & $75.40$ & $78.00$ & $73.03$ & $40.87$ & $45.86$ & $26.70$ \\
    PBPT w/o mutation & $\mathbf{77.73}$ & $77.76$ & $\mathbf{77.54}$ & $70.07$ & $70.47$ & $69.68$ \\
    PBPT & $77.59$ & $\mathbf{78.12}$ & $77.03$ & $\mathbf{76.40}$ & $\mathbf{76.74}$ & $\mathbf{75.79}$ \\
    \bottomrule
  \end{tabular*}
\end{table}

As shown in \cref{tab:enigmata_summary}, PBPT improves mean test accuracy and the test performance of both the highest- and lowest-ranked members under both initial configurations.
Notably, PBPT achieves these gains on Enigmata even under the tuned initialization, where the independent RLVR baseline is already particularly strong and carefully tuned, leaving relatively little headroom for improvement. We observe that the gap between PBPT and RLVR is larger on the other environments.

Under the tuned initialization, PBPT still improves all three measures.
The smaller gap for the highest-ranked member reflects the fact that an occasional well-behaved RLVR trajectory can already reach high performance when initialized from a well-tuned configuration.
PBPT's advantage is therefore most visible in the population as a whole: it raises average performance and improves the lower end of the population while retaining strong top-performing members.

The advantage becomes much larger under the suboptimal initialization.
Here, PBPT substantially outperforms independent RLVR across all three measures, showing that mutation enables the population to move away from a poor initial configuration while selection propagates successful learners.

\subsection{Training with Transferred Configurations}
\label{subsec:res_transfer}

We next study whether PBPT's improvement over RLVR can be explained solely by finding better initial hyperparameters.
On Nemotron, we aggregate the final configurations of the PBPT population into a single fixed configuration, taking the median for real-valued settings, the lower median for integer settings, and a majority vote for Boolean and categorical settings.
We use this configuration to train $8$ independent RLVR learners from the base model, denoted \emph{RLVR (transfer config)}.
We compare these learners with PBPT and the original RLVR baseline.
Each method trains $8$ Qwen3.5-4B learners for the same duration, and
all transfer learners use the same configuration throughout training, differing only in their random seeds.

\begin{figure}[tbp]
  \centering
  \includegraphics[width=\textwidth]{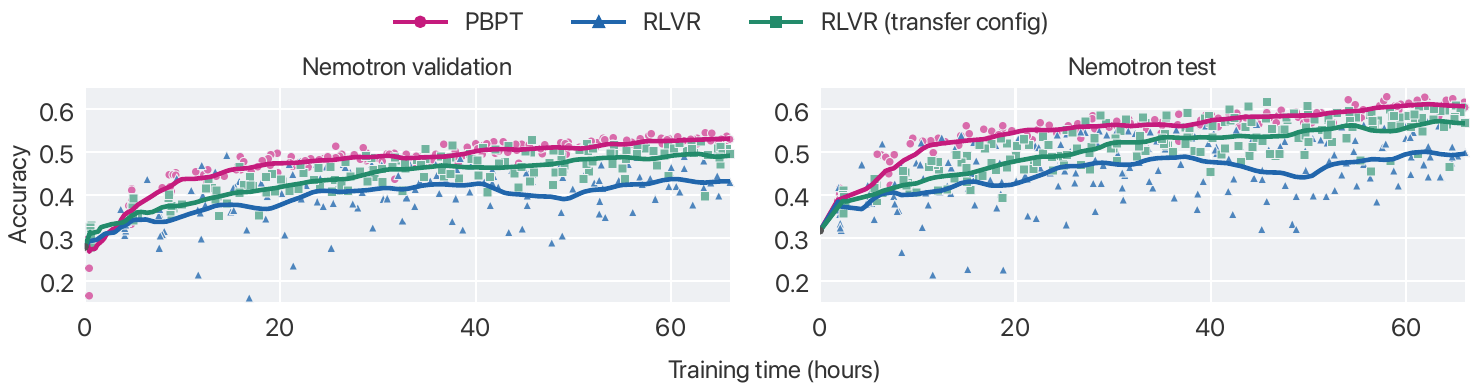}
  \caption{Training with transferred configurations on Nemotron.
  Independent RLVR starts from the base model using a single fixed configuration aggregated from the final PBPT population.
  The axes, points, and lines follow the same convention as in \cref{fig:main_results}.
  The transferred configuration improves mean validation and test performance over the original RLVR baseline, but does not match PBPT.}
  \label{fig:transfer_results}
\end{figure}

As shown in \cref{fig:transfer_results}, RLVR with the aggregated configuration outperforms the original RLVR baseline on both validation and test splits, showing that PBPT's final population provides better initial hyperparameters.
However, these learners still fall behind PBPT, so its gains cannot be explained solely by finding a better starting configuration.
Together with the online-mutation-only ablation in \cref{app:additional_performance}, these results support the importance of transferring successful training states during population optimization.

\begin{figure}[tbp]
  \centering
  \begin{minipage}[b]{0.52\textwidth}
    \begin{subfigure}[b]{\linewidth}
      \includegraphics[width=\linewidth]{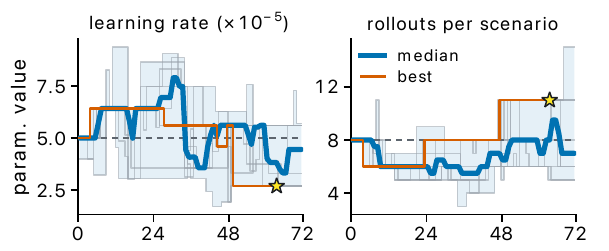}
      \caption{Hyperparameter value evolution during training}
      \label{fig:behavior_knobs}
    \end{subfigure}\\[2pt]
    \begin{subfigure}[b]{\linewidth}
      \includegraphics[width=\linewidth]{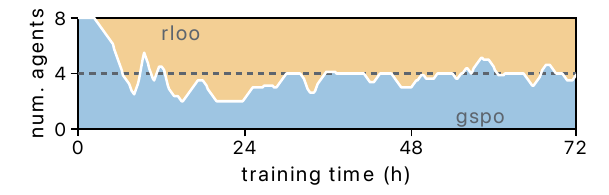}
      \caption{Number of members using each RL loss}
      \label{fig:behavior_loss}
    \end{subfigure}
  \end{minipage}\hfill
  \begin{subfigure}[b]{0.47\textwidth}
    \includegraphics[width=\linewidth]{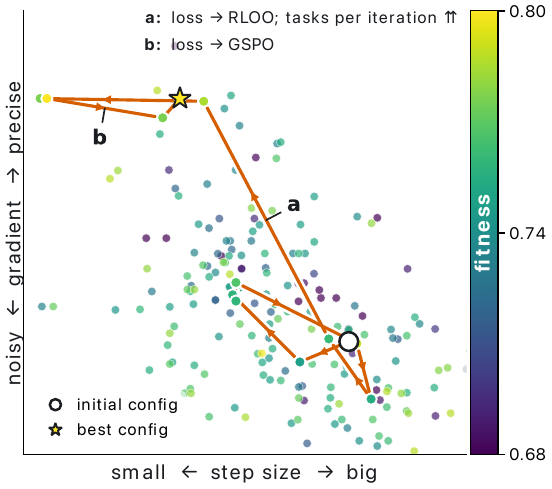}
    \caption{Hyperparameter path to the best checkpoint}
    \label{fig:behavior_hparams}
  \end{subfigure}
  \caption{Evolution of the training configurations in the PBPT population on Enigmata.
  Subfigure~(a) shows the learning rate and rollouts per scenario for individual members in gray, their range as a shaded band, and the population median in blue.
  Dashed lines mark initial values.
  Subfigure~(b) shows the number of members using each RL loss.
  Subfigure~(c) maps explored configurations to two interpretable axes, with points colored by the best fitness recorded under each configuration.
  The orange paths in Subfigures~(a) and~(c) trace the training history leading to the checkpoint with the highest fitness.
  In Subfigure~(c), the circle and star mark the initial and best configurations.}
  \label{fig:behavior_analysis}
\end{figure}

\subsection{PBPT Behavior Analysis}
\label{subsec:res_behavior}

\Cref{fig:behavior_analysis} shows how PBPT explores and adapts training configurations over the course of the Enigmata run from \cref{fig:main_results}.
\Cref{fig:behavior_knobs} shows emergent schedules for selected hyperparameters: the learning rate rises early and then decreases, while the number of rollouts per scenario tends to increase later in training.
The population also continues to explore both GSPO and RLOO throughout training rather than converging to a single loss family (\cref{fig:behavior_loss}).

To visualize PBPT's hyperparameter search, \cref{fig:behavior_hparams} maps configurations onto two axes formed from weighted combinations of hyperparameters (\cref{app:scaling_behavior_vis,tab:hparam-axes}).
The horizontal axis captures optimization aggressiveness and increases with factors such as larger learning rates and looser GSPO clipping.
The vertical axis captures the tradeoff between cheaper, noisier updates and more precise, expensive gradient estimates, increasing with larger batch sizes, more rollouts per scenario, and more on-policy updates as in RLOO.
The best lineage follows a highly non-monotonic path through this space, repeatedly moving between distinct training regimes through mutation.

Overall, PBPT tends to favor larger optimization steps and cheaper, noisier gradients earlier in training for faster wall-time iteration.
It shifts toward more conservative, lower-variance, and more on-policy updates later in training, when useful learning signal becomes sparser.

\FloatBarrier

\section{Discussion and Conclusion}
\label{sec:conclusion}
\label{sec:discussion}

\label{sec:agentic-extensions}

We introduced PBPT, a decentralized framework that integrates online hyperparameter optimization with parallel RLVR post-training by mutating hyperparameters and transferring learned weights from stronger to weaker learners.
We demonstrated that, on reasoning and tool-use benchmarks, PBPT produces stronger agents than compute-matched independent RLVR.
Additionally, PBPT automatically discovers better configurations and recovers from suboptimal ones during training.

While our experiments focused on PBPT with \textit{parametric mutations} (i.e., mutations of numeric hyperparameters and discrete training choices), the PBPT framework can also accommodate \textit{non-parametric mutations}, such as changes to an agent's system prompt or execution harness.
In this setup, an \textit{LLM agent} can make population decisions, selections, and mutations in place of fixed rules.
This is closely related to autonomous experimentation systems such as Autoresearch \citep{karpathy2026autoresearch}, which uses an LLM agent to propose and evaluate changes to training code.
Our preliminary experiments with this extension suggest that current systems remain fragile and highly sensitive to prompting, agent choice, and the representation of candidate modifications.
As agents improve in reliability and long-horizon reasoning, integrating them into PBPT could become a powerful avenue for more autonomous adaptation and an exciting direction for future work.
\ifisarxiv\else
\newpage
\subsection*{AI Use Statement}

We used generative AI tools to assist with manuscript drafting and editing, literature search, code implementation and debugging, figure preparation, and plotting of experimental results.
The human authors developed the core research ideas and algorithmic design.
The authors are responsible for verifying AI-assisted material and for the manuscript, code, and results.

\subsection*{Ethics Statement}

This work studies LLM post-training on reasoning and tool-use benchmarks.
We do not identify ethical concerns specific to PBPT beyond those associated with training and deploying LLMs.

\subsection*{Reproducibility Statement}

To support reproducibility, we describe PBPT in \cref{sec:method} and provide pseudocode in \cref{alg:pbpt,alg:pbpt_operators}.
\Cref{sec:experiments} specifies the benchmarks, data splits, models, evaluation protocol, and training resources.
\Cref{app:experimental_details} provides further details on benchmark modifications, initial RL configurations, reward functions, PBPT search spaces, and implementation.
\fi

\bibliography{references/references}
\bibliographystyle{iclr2027_conference}

\newpage
\appendix
\section{Experimental Details}
\label{app:experimental_details}

\subsection{Benchmarks and Evaluation}
\label{app:benchmark_details}

We describe each benchmark's data, evaluation, and rewards, using the term \emph{scenario} to denote a task instance.
The decoding settings below give the initial values for each method.
Both methods always sample training rollouts at temperature $1$, without top-\textit{p}, top-\textit{k}, or min-\textit{p} filtering.

PBPT may mutate the evaluation temperature without affecting training rollouts.
On Gaia2 and AppWorld, the per-message generation limit is also mutable for training and evaluation.
Validation and test use each member's current settings, whereas independent RLVR keeps these settings fixed.
Within each benchmark, the total evaluation budget per task is fixed across PBPT and independent RLVR checkpoints, regardless of these changes.

\paragraph{Evaluation schedule.}
Each learner saves one checkpoint per training iteration.
Evaluation proceeds in \emph{passes}, and each pass selects one checkpoint.
The selected checkpoint is the newest finished checkpoint that is at least $1$ and at most $20$ iterations (Enigmata, Nemotron) or $10$ iterations (Gaia2, AppWorld) after the previously selected one.
Evaluation therefore skips checkpoints when training outpaces it.
A pass evaluates the selected checkpoint on the validation split if the pass index is a multiple of $p_{\mathrm{val}}$ and on test if it is a multiple of $p_{\mathrm{test}}$.
A pass on which neither split is due evaluates nothing and skips its checkpoint.
The spacing between evaluated checkpoints therefore depends on evaluation speed, and we report measured spacings for each benchmark below.

\paragraph{Compute.}
In all experiments, each PBPT member and each independent RLVR seed runs on one node with $4$ B200 GPUs.
The learner measures its peak memory use and gives the remaining GPU memory to the vLLM inference servers.
Evaluation runs concurrently as a separate process on the same node, using these inference servers.
The same GPUs host policy updates, rollout generation, evaluation, and, on Gaia2, the LLM judges, with no separate evaluation allocations.
Concurrent evaluation slows training iterations and therefore contributes to the training-time clock.

For the $8$-learner comparisons, each method has a nominal training budget of $8\times4\times66=2{,}112$ GPU-hours on Enigmata, Nemotron, and AppWorld, and $8\times4\times90=2{,}880$ GPU-hours on Gaia2.
These budgets exclude overhead outside training time, such as cluster queueing and preemptions.

\paragraph{Curve aggregation and smoothing.}
For each metric, we first construct a population mean over evaluation time.
At each evaluation time appearing in any learner's trajectory, we average the most recent available score from every learner that has already been evaluated.
If a learner stops early, its final score continues to contribute to the mean.

For visualization (see \Cref{fig:main_results,fig:adaptation_results,fig:ablation_results}), we smooth the resulting mean curve with a Gaussian kernel that uses observations both before and after each plotted time.
At training time $t$ hours, the kernel standard deviation is
$0.75\min\{1,\max\{0.05,t/12\}\}$ hours.
We subsample scatter points only for readability, targeting $150$ points per method and panel ($100$ on AppWorld); this does not affect the mean curves.
\Cref{tab:enigmata_summary} instead reports unsmoothed evaluations within its stated averaging window.

\subsubsection{Enigmata}
\label{app:enigmata}

\paragraph{Data.}
Enigmata \citep{DBLP:conf/nips/ChenHYCCDYCLYZW25} contains $36$ puzzle families across $7$ reasoning categories, including Minesweeper, Sudoku, and Countdown.
We train and evaluate on a significantly modified version of Enigmata with bug fixes (see ``Benchmark corrections and improvements'' below).

Each task requires a single response without tool use.
We generate instances using the official Enigmata script and construct the validation set by shuffling the generated data and holding out a subset.
The remaining instances form the training set, and we sample the test set from Enigmata-Eval.
After the corrections described below, the splits comprise $161{,}213$ training, $229$ validation, and $998$ test instances over $35$ families.
Task identifiers are unique across the pooled splits.
We also checked task families for overlap between training and evaluation.
In the crossword family, $84.8\%$ of evaluation clue--answer pairs appeared verbatim in training, contributing to its removal.
No regenerated fifteen-puzzle evaluation board matches or is one move away from a training board.

\paragraph{Evaluation.}
We sample $2$ rollouts per validation task and $1$ per test task.
A rule-based verifier scores each response as $1$ if correct and $0$ otherwise, and we report the mean over all rollouts.
The initial evaluation temperature is $0.6$, with top-\textit{p} set to $0.95$ and top-\textit{k} to $20$.
We do not use multi-token prediction.
We allow $48{,}000$ context tokens and up to $24{,}576$ generated tokens per response.
Both methods use $p_{\mathrm{val}}=2$.
In the main comparison, PBPT uses $p_{\mathrm{test}}=4$ and independent RLVR uses $p_{\mathrm{test}}=6$.
The baseline's evaluation passes also include additional held-out splits not reported here.
In the reported runs, consecutive validated checkpoints are a median of $2$ training iterations ($0.4$ hours) apart for PBPT and $4$ iterations ($0.55$ hours) for independent RLVR.
Consecutive test evaluations are a median of $5$ iterations ($0.8$ hours) apart for PBPT and $14$ iterations ($2.3$ hours) for independent RLVR.
In the suboptimal-configuration experiments, all three methods use $p_{\mathrm{test}}=4$, and validated checkpoints are a median of $2$ to $3$ iterations apart.

\paragraph{Reward.}
The task-specific verifier checks the extracted answer against the reference solution or the puzzle's constraints.
The training reward is $1$ for a correct response and $0$ otherwise, with no length, format, or other penalty terms.
We score only the text after the final \texttt{</think>} tag, which marks the end of the reasoning trace.
A response with an unclosed reasoning trace scores $0$.
The verifier rejects missing or malformed answers with a score of $0$, as well as any response on which it raises an exception.
Responses truncated at the generation limit therefore score $0$, and we keep them as training examples.
We exclude only system-cancelled rollouts from gradient steps.
The system prompt states the remaining token budget.

\begin{figure}[tbp]
  \centering
  \includegraphics[width=\textwidth]{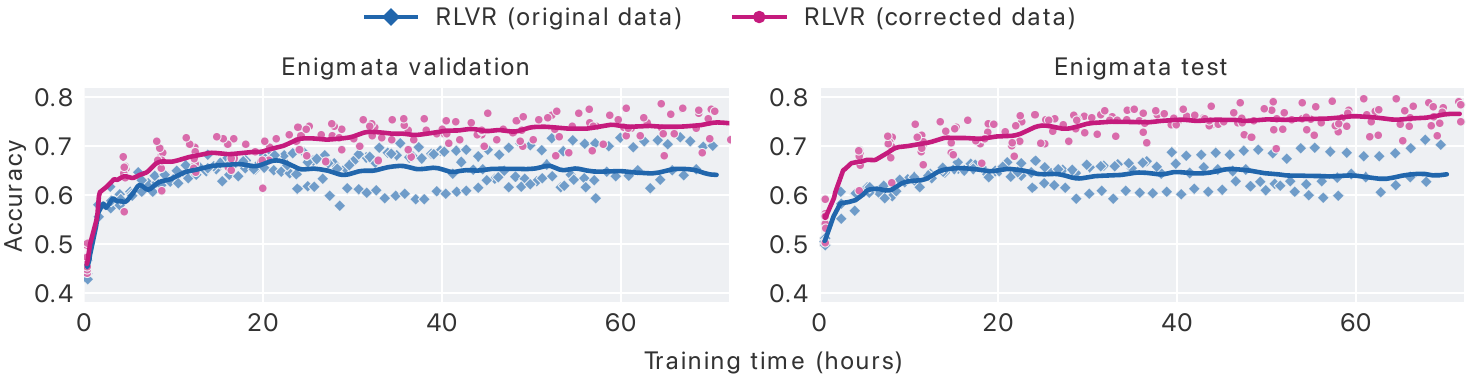}
  \caption{Enigmata results before and after the benchmark corrections of \cref{app:enigmata}.
  We compare independent RLVR learners trained on the original and corrected data.
  We reweight original-data scores to match the corrected splits' family composition, without re-evaluating checkpoints.
  Scatter points show subsampled evaluation results, and solid lines show smoothed means of the latest available score from each learner.
  }
  \label{fig:enigmata_bugfix}
\end{figure}

\paragraph{Benchmark corrections and improvements.}
We corrected defects in the original data and verifiers that affected training rewards or evaluation scores.
\begin{itemize}[leftmargin=1.5em,itemsep=0.2em,topsep=0.3em]
  \item We removed 13,664 duplicate training instances and 498 training and 1 validation function-induction instances whose examples did not determine a unique answer.
  \item We removed the crossword family, whose puzzles were machine-assembled from a pool of clues with fabricated filler letters.
  Counting an entry as a word if its \texttt{wordfreq} Zipf frequency is at least $1.0$, only $2.8\%$ of the training crosswords had solutions in which every entry was a word.
  In addition, $84.8\%$ of the clue--answer pairs in the evaluation splits appeared verbatim in training.
  \item The \emph{fifteen-puzzle} generator's solvability test contained two sign errors, so it admitted $3{,}070$ unsolvable boards among $6{,}024$.
  Always answering ``no solution'' therefore earned about half of the family's score.
  We replaced these boards with boards generated by walking back from the solved state.
  For depths up to $14$, an exhaustive breadth-first search certifies the optimal solution length.
  For deeper boards, iterative deepening A$^*$ verifies the target distance.
  \item We rewrote prompts and regenerated instances for $5$ families whose stated rules did not match their verifiers.
  For \emph{countdown}, we removed a rule that intermediate results be positive integers, which $27.2\%$ of reference solutions violate.
  For \emph{light up}, we added the missing requirement that every empty cell be lit.
  For \emph{magic square}, we stated that all entries are distinct.
  For \emph{minesweeper}, we asked for all determinable cells.
  For \emph{slant}, we corrected the diagonal-symbol descriptions.
  \item We fixed verifier defects in $13$ families.
  For \emph{magic square}, the verifier rejected negative entries, so it rejected $48.8\%$ of its own reference answers.
  For the two \emph{Hamiltonian path} families, the verifier never accepted a ``no path'' answer.
  For \emph{light up} and \emph{natural-language navigation}, the verifier required an exact match to one stored solution although other solutions are valid.
  For \emph{car painting}, the reference answer used a suboptimal heuristic that better solutions can beat.
  For the four \emph{sliding-puzzle} families, the generator never derived solvability from the board.
  For \emph{minesweeper}, the verifier compared answers against an incomplete reference and often rejected correct solutions that identified more mines.
  For \emph{countdown} and \emph{game24}, the verifier accepted operators outside $+,-,\times,\div$ contradicting the task definition and enabling reward hacking.
  The fixed verifiers check the puzzle constraints directly.
  We remove the reasoning trace before verification and assign a score of $0$ to responses that exhaust the generation budget before completing the trace.
\end{itemize}
All other task instances not mentioned above are unchanged.
\Cref{fig:enigmata_bugfix} compares independent RLVR learners trained on the original and corrected Enigmata data.
To make the original-data results more comparable to the corrected split, we recompute their aggregate accuracy from the recorded per-family accuracies, weighting each family by its number of tasks in the corrected split and excluding crosswords (since bug fixes and filtering slightly changed the task distribution).
This only changes how we aggregate the existing family-level results; we do not re-evaluate the original checkpoints on the corrected tasks or verifiers, so this comparison is approximate.

\subsubsection{Nemotron}
\label{app:nemotron}

\paragraph{Data.}
Nemotron \citep{DBLP:journals/corr/abs-2512-20848} is NVIDIA's Nemotron-3-Nano RL training blend of $93{,}244$ instances across $7$ task subsets.
We use the $6$ single-turn subsets: STEM multiple-choice questions, competition mathematics from DAPO-Math-17k and Skywork-OR1, instruction following, JSON generation against a supplied schema, and competitive programming.
The programming tasks contain up to $50$ tests each.
We exclude the multi-turn workplace-assistant subset, as Gaia2 and AppWorld already cover multi-turn tool use.
The released blend omits the mathematics prompts, so we restore the missing $22{,}056$ prompts from the source datasets.
We construct training, validation, and test sets with no shared prompts.
We select validation tasks from instances whose prompts do not occur in the training pool, and we remove test tasks drawn from the training pool together with all copies of their prompts.
The splits contain $81{,}725$ training tasks and $512$ tasks each for validation and test.
Validation and test contain the same number of tasks from each subset.
Each evaluation split contains $160$ multiple-choice, $139$ instruction-following, $97$ Skywork, $43$ JSON, $39$ programming, and $34$ DAPO tasks.

\paragraph{Evaluation.}
We use task-specific rule-based verifiers without an LLM judge.
Multiple-choice answers must match the supplied answer patterns, and a symbolic verifier grades mathematical answers.
Instruction-following responses must satisfy all listed constraints, JSON responses must validate against their supplied schemas, and programming responses must pass all tests.
We score only the text after the final \texttt{</think>} tag, which marks the end of the reasoning trace.
A response with an unclosed reasoning trace scores $0$.
We sample $2$ rollouts per task on both the validation and test splits and report mean accuracy over all rollouts.
A correct response scores $1$, and an incorrect response scores $0$.
Responses that exhaust the generation budget without an answer or cannot be graded by the verifier also score $0$ rather than being excluded.
The initial evaluation temperature is $0.6$, with top-\textit{p} set to $0.95$ and top-\textit{k} to $20$.
We do not use multi-token prediction.
We fix the context and generation limits at $48{,}000$ and $32{,}768$ tokens, respectively.
Validation and test use the member's current evaluation temperature.
After a PBPT action, the evaluation process restarts with the new configuration and evaluates only checkpoints trained after that action.
Population decisions use only validation accuracy as fitness.

We use $p_{\mathrm{val}}=2$ and $p_{\mathrm{test}}=6$ under the evaluation schedule above.
Selection resets the pass counter, so the schedule need not remain periodic across population decisions.
Within the plotted interval, successive validation evaluations are a median of $3$ training iterations apart for PBPT and $4$ for independent RLVR, compared with $11$ and $15$ for test.

\paragraph{Reward.}
We use the same task-specific verifier to assign a binary training reward: $1$ for an accepted response and $0$ otherwise.
Programming tasks receive no partial credit for passing only some tests.
If the verifier cannot grade a response, we assign $0$ and retain the rollout in its group.
There are no additional length or format penalties.
We grade and retain responses truncated at the generation limit, and exclude cancelled rollouts from policy updates.
The system prompt states the generation budget available for the task.

\subsubsection{Gaia2}
\paragraph{Data.}
Gaia2 \citep{DBLP:journals/corr/abs-2602-11964} uses the Mobile environment of Meta's ARE simulator, with email, chat, calendar, and contacts.
Agents complete user requests by interacting with its $101$ APIs.
We use the public validation split, which contains $800$ scenarios across the $5$ capability categories listed in \cref{sec:experiments}.
We split these scenarios into training, validation, and test sets by universe, keeping the underlying application data disjoint.
Universes $21$, $23$, $26$, $27$, $28$, and $30$ provide $458$ training scenarios, universes $24$ and $29$ provide $153$ validation scenarios, and universes $22$ and $25$ provide $189$ test scenarios.
All splits use the same applications and tools.

\paragraph{Evaluation.}
We use Qwen3.5-9B for all LLM judges.
The judge is the base model without the policy's LoRA adapter, served on the learner's own inference servers.
The judges use thinking mode and temperature $0.5$.
Each judge call generates one response of up to $16{,}384$ tokens.
We sample $2$ rollouts per validation scenario and $1$ per test scenario.
We report task goal completion, the fraction of successful rollouts.
The initial policy evaluation temperature is $0.5$, without top-\textit{p} or top-\textit{k} filtering.
Each policy call may initially generate up to $6{,}000$ tokens, and PBPT mutates this limit within $[4{,}000,10{,}000]$.
We do not use multi-token prediction.
Each rollout has a maximum of $200$ steps, matching the original benchmark configuration.
We use $p_{\mathrm{val}}=1$ and $p_{\mathrm{test}}=4$, so every pass includes validation.
Because Gaia2 evaluation is slow, consecutive validated checkpoints are a median of $3$ training iterations (about $1.1$ hours) apart, and test evaluations about $11$ iterations (about $5$ hours) apart, for both methods.

\paragraph{Judging procedure.}
We use ARE's per-event judge to compare the agent's actions against the task's reference (oracle) events.
For each turn, the number of calls to each tool must match the oracle's, except that the agent may send one additional message to the user.
The judge then matches each oracle event to a distinct agent event.
A match requires the agent event to fall within a timing window relative to the events it depends on and to pass deterministic checks on its arguments.
It must also pass LLM-based checks of properties such as content, plausibility, and tone.
The judge must also match each oracle event on which it depends to an earlier agent event.
A rollout succeeds if and only if the agent matches every oracle event of every turn.
Each LLM-based check receives the agent's and the oracle's action arguments with task context and returns a Boolean verdict with a rationale.
If no verdict parses within $3$ attempts, the judge scores the rollout as a failure and we retain it for training.

We handle policy-side timeouts the same way: a generation stall of $300$ seconds, or a total generation time or rollout wall-clock time of $1{,}800$ seconds, ends the rollout as a failure.
Tool errors returned by the simulated applications are ordinary observations.
We discard without a reward only rollouts affected by infrastructure failures, namely judge connection errors or timeouts and rollout processes killed after $2{,}400$ seconds.

\paragraph{Reward.}
The training reward is binary: $1$ for success and $0$ for failure.
Training and evaluation use the same judging procedure and settings.
They differ only in the policy's sampling temperature, the number of rollouts per scenario, and the number of concurrent scenario runners.

\paragraph{Versions and modifications.}
We use a fork of ARE and the public Gaia2 scenarios released on Hugging Face as \texttt{meta-agents-research-environments/gaia2}.
Relative to upstream, our fork makes the following changes:
\begin{itemize}[leftmargin=1.5em,itemsep=0.2em,topsep=0.3em]
  \item LLM judges return schema-validated tool-call verdicts with $1$ rather than $3$ votes to reduce cost.
  \item We recover verdicts from judge text and strip reasoning traces before subtask extraction.
  \item LLM-based checks run in parallel where possible for performance reasons.
  \item The agent uses native tool calling rather than a text ReAct loop.
  \item The agent's system prompt clarifies reporting to the user and waiting for notifications.
  \item We truncate observations and add wall-clock caps.
  \item We classify judge infrastructure failures separately from judge verdicts.
  \item We apply the configured step limit.
\end{itemize}

\subsubsection{AppWorld}
\paragraph{Data.}
AppWorld \citep{DBLP:conf/acl/TrivediKHMDLGSB24} evaluates personal-assistant agents in a simulator with $10$ applications, including Amazon, Spotify, Venmo, and Gmail.
Each task consists of a user request, such as transferring money or gathering listening statistics.
The agent fulfills the request through successive Python code executions that call application APIs.
We use version \texttt{0.1.3} of the \texttt{appworld} package.
We use the \texttt{train} split with $90$ tasks for training.
For validation, we use $55$ of the $57$ tasks in the \texttt{dev} split, excluding tasks \texttt{3ab5b8b\_3} and \texttt{37a8675\_1} because of bugs in their task definitions.
We report results separately on \texttt{test\_normal} with $168$ tasks and \texttt{test\_challenge} with $417$ tasks.

\paragraph{Evaluation.}
We sample $2$ rollouts per validation task and $1$ per task in each test split.
A task succeeds only if all required checks on the response and final application state pass.
We report task goal completion, the fraction of successful rollouts.
The initial evaluation temperature is $0.1$, without top-\textit{p} or top-\textit{k} filtering, with a context limit of $65{,}536$ tokens and a generation limit of $8{,}192$ tokens per policy call.
We allow up to $30$ interactions with the Python interpreter during training and $50$ during evaluation.
We use $p_{\mathrm{val}}=1$ and $p_{\mathrm{test}}=3$, so every pass includes validation.
Consecutive validated checkpoints are a median of $1$ training iteration ($0.45$ hours for PBPT and $0.6$ hours for independent RLVR) apart, and test evaluations $3$ to $4$ iterations (about $1.5$ to $1.8$ hours) apart.
There is no wall-clock limit on the policy.
The environment returns any error in the agent's own code to the agent as an observation.
Infrastructure errors, such as a failed or timed-out call to the AppWorld server ($100$ seconds per call), restart the server and rerun the episode from scratch, up to $5$ attempts.
If all attempts fail, we abort the training iteration or evaluation pass rather than silently drop the rollout.

\paragraph{Reward.}
At the end of each rollout, the training reward is the fraction of task-specific checks passed.
For example, passing $3$ of $4$ checks earns $0.75$ but does not count as success in evaluation.
We also score rollouts that exhaust interaction or context budgets, with no extra penalty.



\FloatBarrier
\subsection{RL Configurations and Search Spaces}
\label{app:initial_configs}


Within each comparison, all methods use the same configuration before initial mutation.
The initial values in the tables below are launch values, identical for every member and seed.
PBPT applies an initial mutation to $5$ of its $8$ members in the main experiments and to $6$ of $8$ in the suboptimal-configuration experiment.
The initial mutation uses the same mutation probability and scaling range as later mutations.
Independent RLVR and PBPT without mutation do not apply an initial mutation.

\subsubsection{Main Experiments}

\paragraph{Shared settings.}
\Cref{tab:reasoning-ranges,tab:agentic-ranges} give the initial configurations and PBPT search ranges for the reasoning and agentic tool-use benchmarks, respectively.
PBPT uses $p_{\mathrm{mut}}=0.2$ and numeric scaling factors $\mu\sim\operatorname{Uniform}[1.1,1.5]$.
\Cref{app:implementation_details} describes the mutation operator.

\begin{table}[t]
\centering
\caption{Initial RL configurations and PBPT search ranges for the reasoning benchmarks.
Values precede initial mutation, ``fixed'' means not mutated, and ``---'' denotes inapplicable settings.}
\label{tab:reasoning-ranges}
\small
\setlength{\tabcolsep}{3.4pt}
\begin{tabular*}{\textwidth}{@{\extracolsep{\fill}}lcccc@{}}
\toprule
& \multicolumn{2}{c}{Enigmata ($4$B)} & \multicolumn{2}{c}{Nemotron ($4$B)} \\
\cmidrule(lr){2-3}\cmidrule(l){4-5}
Hyperparameter & Initial value & Search range & Initial value & Search range \\
\midrule
\multicolumn{5}{@{}l}{\textit{Optimization}} \\
Learning rate & $5\times10^{-5}$ & $[10^{-8}, 10^{-3}]$ & $5\times10^{-5}$ & $[10^{-8}, 10^{-3}]$ \\
Weight decay & $0.01$ & $[10^{-5}, 0.02]$ & $0.01$ & $[10^{-5}, 0.02]$ \\
\multicolumn{5}{@{}l}{\textit{RL core}} \\
RL objective & GSPO & \{RLOO, GSPO\} & GSPO & \{RLOO, GSPO\} \\
Epochs per iteration & $2$ & $[1, 4]$ & $2$ & $[1, 4]$ \\
Rollouts per scenario & $8$ & $[3, 16]$ & $8$ & $[3, 16]$ \\
Scenarios per iteration & $40$ & $[8, 96]$ & $40$ & $[8, 96]$ \\
Minibatch size & $64$ & $[8, 256]$ & $64$ & $[8, 256]$ \\
Max gradient norm & $0.1$ & $[10^{-6}, 1]$ & $0.1$ & $[10^{-6}, 1]$ \\
\multicolumn{5}{@{}l}{\textit{Trust region}} \\
Lower clipping parameter $\epsilon_{\mathrm{low}}$ & $3\times10^{-4}$ & $[2\times10^{-4}, 0.01]$ & $3\times10^{-4}$ & $[2\times10^{-4}, 0.01]$ \\
Upper clipping parameter $\epsilon_{\mathrm{high}}$ & $4\times10^{-4}$ & $[2\times10^{-4}, 0.01]$ & $4\times10^{-4}$ & $[2\times10^{-4}, 0.01]$ \\
Truncated importance sampling $c$ & $2$ & $[1, 10]$ & $2$ & $[1, 10]$ \\
Clip log-prob grads at $-\epsilon$ & on & \{on, off\} & on & \{on, off\} \\
\multicolumn{5}{@{}l}{\textit{Entropy}} \\
REPO-R rescaling & off & \{on, off\} & off & \{on, off\} \\
REPO-R $\beta_{\max}$ & $0.05$ & $[0.01, 0.2]$ & $0.05$ & $[0.01, 0.2]$ \\
\multicolumn{5}{@{}l}{\textit{Rollout sampling}} \\
Non-zero-advantage fraction & $0.25$ & $[0.1, 0.5]$ & $0.25$ & $[0.1, 0.5]$ \\
Rollouts fraction & $0.9$ & $[0.1, 0.99]$ & $0.9$ & $[0.1, 0.99]$ \\
Rollouts-per-scenario fraction & $0.66$ & $[0.1, 0.99]$ & $0.66$ & $[0.1, 0.99]$ \\
\multicolumn{5}{@{}l}{\textit{Scenario replay}} \\
Scenario replay & off & fixed & on & \{on, off\} \\
Replay fraction & --- & --- & $0.3$ & $[0.05, 0.9]$ \\
Replay max fail streak & --- & --- & $5$ & $[2, 20]$ \\
\multicolumn{5}{@{}l}{\textit{Evaluation}} \\
Evaluation temperature & $0.6$ & $[0.05, 0.75]$ & $0.6$ & $[0.05, 0.75]$ \\
\multicolumn{5}{@{}l}{\textit{Concurrency}} \\
Training scenario runners & $80$ & $[20, 200]$ & $80$ & fixed \\
Evaluation scenario runners & $200$ & $[50, 400]$ & $200$ & fixed \\
\multicolumn{5}{@{}l}{\textit{Generation budget}} \\
Maximum training sequence length & $48{,}000$ & fixed & $48{,}000$ & fixed \\
Maximum generated tokens & $24{,}576$ & fixed & $32{,}768$ & fixed \\
\multicolumn{5}{@{}l}{\textit{Task curriculum}} \\
Per-family sampling weight & $1.0$ (unset) & $[0.05, 20]$ & --- & --- \\
\bottomrule
\end{tabular*}
\end{table}

\begin{table}[t]
\centering
\caption{Initial RL configurations and PBPT search ranges for the agentic tool-use benchmarks.
Values precede initial mutation, and ``fixed'' means not mutated.}
\label{tab:agentic-ranges}
\small
\setlength{\tabcolsep}{3.4pt}
\begin{tabular*}{\textwidth}{@{\extracolsep{\fill}}lcccc@{}}
\toprule
& \multicolumn{2}{c}{Gaia2 ($9$B)} & \multicolumn{2}{c}{AppWorld ($4$B)} \\
\cmidrule(lr){2-3}\cmidrule(l){4-5}
Hyperparameter & Initial value & Search range & Initial value & Search range \\
\midrule
\multicolumn{5}{@{}l}{\textit{Optimization}} \\
Learning rate & $5\times10^{-5}$ & $[10^{-8}, 10^{-3}]$ & $5\times10^{-5}$ & $[10^{-8}, 10^{-3}]$ \\
Weight decay & $0.01$ & $[10^{-5}, 0.02]$ & $0.01$ & $[10^{-5}, 0.02]$ \\
\multicolumn{5}{@{}l}{\textit{RL core}} \\
RL objective & GSPO & \{RLOO, GSPO\} & GSPO & \{RLOO, GSPO\} \\
Epochs per iteration & $2$ & $[1, 4]$ & $2$ & $[1, 4]$ \\
Rollouts per scenario & $6$ & $[3, 16]$ & $6$ & $[3, 16]$ \\
Scenarios per iteration & $32$ & $[8, 96]$ & $64$ & $[8, 96]$ \\
Minibatch size & $64$ & $[8, 256]$ & $128$ & $[8, 256]$ \\
Max gradient norm & $0.1$ & $[10^{-6}, 1]$ & $0.1$ & $[10^{-6}, 1]$ \\
\multicolumn{5}{@{}l}{\textit{Trust region}} \\
Lower clipping parameter $\epsilon_{\mathrm{low}}$ & $3\times10^{-4}$ & $[2\times10^{-4}, 0.01]$ & $3\times10^{-4}$ & $[2\times10^{-4}, 0.01]$ \\
Upper clipping parameter $\epsilon_{\mathrm{high}}$ & $4\times10^{-4}$ & $[2\times10^{-4}, 0.01]$ & $4\times10^{-4}$ & $[2\times10^{-4}, 0.01]$ \\
Truncated importance sampling $c$ & $2$ & $[1, 10]$ & $2$ & $[1, 10]$ \\
Clip log-prob grads at $-\epsilon$ & on & \{on, off\} & on & \{on, off\} \\
\multicolumn{5}{@{}l}{\textit{Entropy}} \\
REPO-R rescaling & off & \{on, off\} & off & \{on, off\} \\
REPO-R $\beta_{\max}$ & $0.05$ & $[0.01, 0.2]$ & $0.05$ & $[0.01, 0.2]$ \\
\multicolumn{5}{@{}l}{\textit{Rollout sampling}} \\
Non-zero-advantage fraction & $0.25$ & $[0.1, 0.5]$ & $0.25$ & $[0.1, 0.5]$ \\
Rollouts fraction & $0.9$ & $[0.1, 0.99]$ & $0.9$ & $[0.1, 0.99]$ \\
Rollouts-per-scenario fraction & $0.66$ & $[0.1, 0.99]$ & $0.66$ & $[0.1, 0.99]$ \\
\multicolumn{5}{@{}l}{\textit{Scenario replay}} \\
Scenario replay & on & \{on, off\} & on & \{on, off\} \\
Replay fraction & $0.3$ & $[0.05, 0.9]$ & $0.3$ & $[0.05, 0.9]$ \\
Replay max fail streak & $5$ & $[2, 20]$ & $5$ & $[2, 20]$ \\
\multicolumn{5}{@{}l}{\textit{Evaluation}} \\
Evaluation temperature & $0.5$ & $[0.1, 0.75]$ & $0.1$ & $[0.05, 0.75]$ \\
\multicolumn{5}{@{}l}{\textit{Concurrency}} \\
Training scenario runners & $16$ & fixed & $24$ & fixed \\
Evaluation scenario runners & $36$ & fixed & $64$ & fixed \\
\multicolumn{5}{@{}l}{\textit{Generation budget}} \\
Maximum training sequence length & $80{,}000$ & fixed & $65{,}536$ & $[48{,}000, 65{,}536]$ \\
Maximum generated tokens & $6{,}000$ & $[4{,}000, 10{,}000]$ & $8{,}192$ & $[4{,}000, 9{,}000]$ \\
\bottomrule
\end{tabular*}
\end{table}

\paragraph{Benchmark-specific settings.}
Enigmata disables scenario replay and includes $35$ mutable task-family sampling weights and $2$ mutable scenario-runner counts.
The other benchmarks initially enable scenario replay and do not mutate task-family weights or scenario-runner counts.

\paragraph{Fixed RL settings.}
All experiments use AdamW with $(\beta_1,\beta_2)=(0.9,0.95)$ and $\epsilon=10^{-15}$.
The learning rate is constant between PBPT actions, without warm-up or decay.
Training uses bfloat16 and LoRA adapters with rank $8$, scaling factor $16$, and no dropout on all attention, feed-forward, and linear-attention projections.
We clip gradients to the configured maximum norm.
Advantages use a leave-one-out baseline without normalization.
There is no Kullback--Leibler (KL) penalty and no entropy bonus.
Under GSPO, each iteration runs the configured number of epochs over minibatches, with sequence-level clipping to $[1-\epsilon_{\mathrm{low}},1+\epsilon_{\mathrm{high}}]$.
Under RLOO, training is strictly on-policy with a single epoch, a single optimizer step, and no GSPO clipping.
In that case, the stored epoch count, minibatch size, and clipping bounds have no effect.
Truncated importance sampling, used under both objectives, corrects for numerical differences between the vLLM sampler and the trainer.
Let $p_\psi$ denote the token probabilities reported by the vLLM sampler, and let $p_{\theta_{\mathrm{old}}}$ denote the probabilities recomputed by the trainer with its parameters at the start of the iteration.
We multiply each output token $x_t$'s loss term by the fixed weight $\min\{p_{\theta_{\mathrm{old}}}(x_t\mid\mathbf{c},\mathbf{x}_{<t})/p_\psi(x_t\mid\mathbf{c},\mathbf{x}_{<t}),\,c\}$.
GSPO computes its sequence-level importance ratio relative to $p_{\theta_{\mathrm{old}}}$ rather than $p_\psi$.
Nemotron uses fully sharded data parallelism across $4$ GPUs, and updates exclude rollouts with absolute advantage below $0.01$.
Nemotron aborts iterations with absolute estimated KL divergence above $20$.
Nemotron uses no frequency penalty and a replay buffer of up to $20{,}000$ scenarios.

\paragraph{Implementation-specific parameters.}
The \emph{REPO-R} entropy regularizer \citep{petrenko2026entropypreserving} replaces each rollout's leave-one-out advantage $A$ with a per-token advantage that depends on the detached token log-probability $\log p_t$.
If $A>0$, then $A_t=\max\{0,A(1-\beta\log p_t)\}$, and if $A<0$, then $A_t=\min\{0,A(1+\beta\log p_t)\}$.
Starting from $\beta=10^{-3}$, the regularizer doubles or halves $\beta$ once per iteration to steer the estimated token entropy toward its value at iteration $0$.
Its sign changes when $|\beta|$ falls below $10^{-4}$, and $|\beta|\le\beta_{\max}$.
$\beta_{\max}$ has no effect while REPO-R is off.

\emph{Log-probability gradient clipping} stops gradients through token log-probabilities of at least $-\epsilon$, where $\epsilon$ is the float32 machine epsilon.
This works around numerical issues for near-certain tokens.

The \emph{rollout collection} target, $\mathrm{rollouts}$, equals scenarios per iteration times rollouts per scenario.
A scenario's group is eligible for collection once at least $\max\{2,\mathrm{round}(f_{\mathrm{rps}}\cdot\text{rollouts per scenario})\}$ of its rollouts complete, where $f_{\mathrm{rps}}$ is the rollouts-per-scenario fraction.
Collection stops once eligible groups contain at least $f_{\mathrm{r}}\cdot\mathrm{rollouts}$ completed rollouts, where $f_{\mathrm{r}}$ is the rollouts fraction.
At least $f_{\mathrm{nz}}\cdot\mathrm{rollouts}$ of these must come from groups with non-zero advantage, where $f_{\mathrm{nz}}$ is the non-zero-advantage fraction.
The collector then cancels unfinished rollouts.
Each completed group with zero advantages triggers sampling of a replacement scenario, up to $10\cdot\mathrm{rollouts}$ collected rollouts.
Updates exclude zero-advantage rollouts.

\emph{Scenario replay} stores task instances whose last group contains both successful and failed rollouts.
Each scenario draw comes from the buffer with probability equal to the replay fraction.
Instances leave after a fully successful group or the maximum failure streak, measured in consecutive fully failed groups.
On AppWorld, a rollout counts as solved only if all checks pass.

The \emph{scenario-runner} counts are the numbers of concurrent rollout workers.
For training, the count is per learner process, with $4$ processes per learner.
For evaluation, it is the total count.
All workers share the learner's vLLM servers.

Enigmata uses \emph{task-family sampling weights} $w_f$ to draw family $f$ with probability $w_f/\sum_g w_g$, then samples an instance uniformly within that family.
Setting all weights to $1$ samples families uniformly, not individual instances.

\subsubsection{Adaptation from Suboptimal RL Configurations}
\label{app:suboptimal_config}

\begin{table}[htbp]
\centering
\caption{
Initial RL configuration and PBPT search ranges for the suboptimal-configuration experiments on Enigmata in \cref{subsec:res_adaptation,subsec:res_ablation}.
The context limit ($48{,}000$ tokens) and generation limit ($24{,}576$ tokens) are fixed.
``Final'' summarizes the PBPT population at the end of training.
For numeric hyperparameters, it gives the median over the values recorded during each member's last $100$ training iterations, pooled across the $8$ members.
For Boolean settings, it gives the number of members ending with the setting on.
``Mixed'' indicates that members ended with different RL objectives ($6$ RLOO, $2$ GSPO).
Under RLOO, the epoch count, minibatch size, and clipping bounds are inactive.
The per-family weights give the median and the $10$th--$90$th percentiles over the final weights of all $35$ families and $8$ members.}
\label{tab:hcap-hparams}
\small
\begin{tabular}{lccc}
\toprule
Hyperparameter & Initial value & Search range & Final \\
\midrule
\multicolumn{4}{l}{\textit{Optimization}} \\
Learning rate & $10^{-4}$ & $[10^{-8}, 10^{-3}]$ & $2.7\times10^{-5}$ \\
Weight decay & $2.5\times10^{-3}$ & $[10^{-5}, 0.02]$ & $3.2\times10^{-3}$ \\
\multicolumn{4}{l}{\textit{RL core}} \\
RL objective & GSPO & \{RLOO, GSPO\} & mixed \\
Epochs per iteration & $4$ & $[1, 4]$ & $3$ \\
Rollouts per scenario & $4$ & $[3, 16]$ & $10$ \\
Scenarios per iteration & $48$ & $[8, 96]$ & $40$ \\
Minibatch size & $24$ & $[8, 256]$ & $80$ \\
Max gradient norm & $0.5$ & $[10^{-6}, 1]$ & $0.223$ \\
\multicolumn{4}{l}{\textit{Trust region}} \\
Lower clipping parameter $\epsilon_{\mathrm{low}}$ & $2\times10^{-3}$ & $[2\times10^{-4}, 0.01]$ & $1.5\times10^{-3}$ \\
Upper clipping parameter $\epsilon_{\mathrm{high}}$ & $2\times10^{-3}$ & $[2\times10^{-4}, 0.01]$ & $4.2\times10^{-3}$ \\
Truncated importance sampling $c$ & $6$ & $[1, 10]$ & $1.28$ \\
Clip log-prob grads at $-\epsilon$ & on & \{on, off\} & $2/8$ on \\
\multicolumn{4}{l}{\textit{Entropy}} \\
REPO-R rescaling & off & \{on, off\} & $4/8$ on \\
REPO-R $\beta_{\max}$ & $0.05$ & $[0.01, 0.2]$ & $0.196$ \\
\multicolumn{4}{l}{\textit{Rollout sampling}} \\
Non-zero-advantage fraction & $0.12$ & $[0.1, 0.5]$ & $0.149$ \\
Rollouts fraction & $0.98$ & $[0.1, 0.99]$ & $0.985$ \\
Rollouts-per-scenario fraction & $0.85$ & $[0.1, 0.99]$ & $0.26$ \\
\multicolumn{4}{l}{\textit{Evaluation}} \\
Evaluation temperature & $0.1$ & $[0.05, 0.75]$ & $0.0574$ \\
\multicolumn{4}{l}{\textit{Concurrency}} \\
Training scenario runners & $30$ & $[20, 200]$ & $30$ \\
Evaluation scenario runners & $200$ & $[50, 400]$ & $184$ \\
\multicolumn{4}{l}{\textit{Task curriculum}} \\
Per-family sampling weight ($35$ families) & $1.0$ (unset) & $[0.05, 20]$ & $1.26$ $[0.14, 5.28]$ \\
\bottomrule
\end{tabular}
\end{table}

\Cref{tab:hcap-hparams} gives the suboptimal initial configuration used in \cref{subsec:res_adaptation,subsec:res_ablation}.
We choose these values manually within PBPT's search ranges to represent plausible choices before careful hyperparameter tuning.
All methods start from this configuration, and only PBPT applies an initial mutation as described above.
For these experiments, PBPT uses $p_{\mathrm{mut}}=0.3$ and numeric scaling factors $\mu\sim\operatorname{Uniform}[1.1,2.5]$.
PBPT without mutation uses the same selection rule but no mutation during training.
All other population-decision settings match the main experiments (\cref{app:implementation_details}).

\clearpage
\subsection{PBPT Implementation Details}
\label{app:search_spaces}
\label{app:implementation_details}

Unless stated otherwise, these settings apply to all benchmarks and experiments.
\Cref{alg:pbpt_operators} gives the decision, crossover, and mutation operators used in \cref{alg:pbpt}.
Here $\operatorname{top}_m$ and $\operatorname{bottom}_m$ return up to $m$ highest- and lowest-fitness records in a set (with at least $1$ when non-empty), and $\operatorname{Std}$ denotes the standard deviation of fitnesses in $\mathcal{C}$.

\begin{algorithm}[t]
\caption{PBPT Decision, Crossover, and Mutation Operators}
\label{alg:pbpt_operators}
\begin{algorithmic}[1]
\Function{Decide}{i,\mathcal{C},\mathcal{S}}
    \If{local progress or peer evidence is insufficient}
        \State \Return $\left(\act{keep},\varnothing\right)$
    \EndIf
    \State $\mathcal{W}\gets\operatorname{top}_{\lceil 0.3|\mathcal{C}|\rceil}\left(\mathcal{C}\right)$ \Comment{Winner set (top 30\%)}
    \State $\mathcal{L}\gets\operatorname{bottom}_{\lceil 0.25|\mathcal{C}|\rceil}\left(\mathcal{C}\right)$ \Comment{Loser set (bottom 25\%)}
    \If{$\mathrm{fitness}_i\ge\min_{\mathcal{C}_j\in\mathcal{W}}\mathrm{fitness}_j$} \Comment{Continue training all winners}
        \State \Return $\left(\act{keep},\varnothing\right)$
    \EndIf
    \If{$\mathcal{C}_i\in\mathcal{L}$}
        \State $\mathrm{best}\gets\max_{\mathcal{C}_j\in\mathcal{C}}\mathrm{fitness}_j$
        \State $\mathrm{spread}\gets\operatorname{Std}_{\mathcal{C}_j\in\mathcal{C}}\left(\mathrm{fitness}_j\right)$
        \State $\tau_{\mathrm{sel}}\gets\max\left\{\mathrm{spread},0.05\left|\mathrm{best}\right|\right\}$ \Comment{Selection threshold}
        \State $\tau_{\mathrm{mut}}\gets\max\left\{0.1\mathrm{spread},0.02\left|\mathrm{best}\right|\right\}$ \Comment{Mutation threshold}
        \If{$\mathrm{best}-\mathrm{fitness}_i>\tau_{\mathrm{sel}}$} \Comment{Large fitness gap: select donor and mutate}
            \State $\mathcal{D}\gets\mathcal{W}\cup\operatorname{top}_{10}\left(\left\{\mathcal{S}_{j,\hat{k}}\in\mathcal{S}\setminus\mathcal{W}\;\middle|\;\mathrm{fitness}_{j,\hat{k}}\ge\mathrm{best},\,\mathrm{time}_{j,\hat{k}}\le\mathrm{time}_i\right\}\right)$
            \State $\mathrm{donor}\sim\operatorname{Uniform}\left(\mathcal{D}\right)$
            \State \Return $\left(\act{replace},\mathrm{donor}\right)$
        \ElsIf{$\mathrm{best}-\mathrm{fitness}_i>\tau_{\mathrm{mut}}$} \Comment{Moderate fitness gap: mutate only}
            \State \Return $\left(\act{perturb},\varnothing\right)$
        \EndIf
    \EndIf
    \State \Return $\left(\act{keep},\varnothing\right)$
\EndFunction
\Statex
\Function{Crossover}{h,h_{\mathrm{donor}}}
    \For{hyperparameter $h_q\in h$}
        \State $h_q\sim\operatorname{Uniform}\left(\left\{h_q,h_{\mathrm{donor},q}\right\}\right)$
    \EndFor
\EndFunction
\Statex
\Function{Mutate}{h}
    \For{mutable hyperparameter $h_q\in h$}
        \If{$r \sim \operatorname{Uniform}[0,1] < p_{\mathrm{mut}}$} \Comment{With probability $p_{\mathrm{mut}}$}
            \State $\mu\sim\operatorname{Uniform}[\mu_{\min},\mu_{\max}]$ and $s\sim\operatorname{Uniform}\left(\left\{-1,+1\right\}\right)$
            \If{$\mathcal{H}_q\subseteq[0,1]$}
                \State $h_q\gets h_q\cdot\mu^s$ if $h_q\le 0.5$, else $1-\left(1-h_q\right)\cdot\mu^s$
                \State $h_q\gets\operatorname{clip}\left(h_q,\,\mathcal{H}_q^{\min},\,\mathcal{H}_q^{\max}\right)$ \Comment{Clip to $\mathcal{H}_q$}
            \ElsIf{$h_q$ is numeric}
                \State $h_q\gets\operatorname{clip}\left(h_q\cdot\mu^s,\,\mathcal{H}_q^{\min},\,\mathcal{H}_q^{\max}\right)$ \Comment{Clip to $\mathcal{H}_q$ (round if integer)}
            \Else \Comment{Boolean or categorical}
                \State $h_q\sim\operatorname{Uniform}\left(\mathcal{H}_q\setminus\left\{h_q\right\}\right)$
            \EndIf
        \EndIf
    \EndFor
\EndFunction
\end{algorithmic}
\end{algorithm}

\paragraph{Clocks.}
Each member has two clocks.
The \emph{runner process time} ($\mathrm{time}_{i,k}$ in \cref{sec:method}) includes training, evaluation, and restarts but pauses while the job is suspended.
It determines compute eligibility, the warm-up and cooldown intervals, and the per-member budget in \cref{alg:pbpt}.
The \emph{training time} accumulates only the wall-clock time of training iterations (rollout collection and policy updates) and is the x-axis of the experimental figures.
Selection preserves the recipient's clocks and iteration counter.
Iteration-based checks use the evaluated checkpoint's training-iteration index.

\paragraph{Peer comparisons.}
A new validation evaluation triggers a population decision for member $i$.
For each peer $j$, member $i$ selects the most recent record $\mathcal{S}_{j,\hat{k}}\in\mathcal{S}$ with $\mathrm{time}_{j,\hat{k}}\le\mathrm{time}_{i,k}$, as described in \cref{sec:method}.
Each record also stores a wall-clock publication timestamp for recency checks.
Member $i$ excludes peer $j$ from $\mathcal{C}_{i,k}$ if $\mathrm{time}_{j,\hat{k}}$ lags $\mathrm{time}_{i,k}$ by more than $2$ hours and peer $j$ published $\mathcal{S}_{j,\hat{k}}$ more than $2$ hours ago in wall-clock time.

\paragraph{Decision eligibility.}
For a population of $n$ members, selection and mutation require $|\mathcal{C}_{i,k}|\ge\lfloor n/2\rfloor$ (including member $i$'s own record), at least $30$ minutes of runner process time, and an evaluated checkpoint from training iteration $3$ or later.
After \act{replace} or a mutation that changes $h_i$, further selection or mutation requires a cooldown of at least $40$ more minutes of runner process time and $2$ more training iterations ($1$ on Gaia2).
Initial mutation also starts this cooldown, so initially mutated members become eligible no earlier than $40$ minutes, compared with $30$ minutes for the others.
Mutations that leave $h_i$ unchanged do not restart the cooldown.

\paragraph{Population decisions.}
With all $8$ members represented in $\mathcal{C}$, the winner set $\mathcal{W}$ and loser set $\mathcal{L}$ in \cref{alg:pbpt_operators} contain $3$ and $2$ records, respectively (for non-empty $\mathcal{C}$, both sets always contain at least $1$ record, and \textsc{Decide} returns \act{keep} if $\mathrm{fitness}_i$ ties with a record in $\mathcal{W}$).

\paragraph{Donor selection and state transfer.}
Every published record $\mathcal{S}_{j,\hat{k}}$ remains in $\mathcal{S}$ for the whole run.
For \act{replace}, the donor pool $\mathcal{D}$ combines the current winner set $\mathcal{W}\subseteq\mathcal{C}$ with up to $10$ historical records $\mathcal{S}_{j,\hat{k}}\in\mathcal{S}\setminus\mathcal{W}$ from any member (including $i$) that satisfy $\mathrm{fitness}_{j,\hat{k}}\ge\mathrm{best}$ and $\mathrm{time}_{j,\hat{k}}\le\mathrm{time}_i$, ordered by descending fitness.
PBPT samples $\mathrm{donor}\sim\operatorname{Uniform}\left(\mathcal{D}\right)$.
Recipient $i$ copies the donor's policy parameters $\theta_{\mathrm{donor}}$ and optimizer state $\omega_{\mathrm{donor}}$, applies \textsc{Crossover} and \textsc{Mutate} in-place to $h_i$, and resumes training as in \cref{alg:pbpt}.

After a training or evaluation crash, or after no new evaluation arrives for $2.5$ hours ($4$ hours on Gaia2), the runner forces member $i$ to take \act{replace} or \act{perturb}, outside the rule above.

\paragraph{Configuration mutation.}
\Cref{app:initial_configs} specifies each experiment's mutation probability $p_{\mathrm{mut}}$ and numeric scaling bounds $\mu_{\min}$ and $\mu_{\max}$.
\textsc{Mutate} perturbs each mutable hyperparameter $h_q\in h$ independently with probability $p_{\mathrm{mut}}$.
For Boolean and categorical hyperparameters (including the RL objective in $\{\text{RLOO},\text{GSPO}\}$), \textsc{Mutate} samples $h_q\sim\operatorname{Uniform}\left(\mathcal{H}_q\setminus\left\{h_q\right\}\right)$.
For a numeric hyperparameter $h_q$, \textsc{Mutate} samples $\mu\sim\operatorname{Uniform}[\mu_{\min},\mu_{\max}]$ and $s\sim\operatorname{Uniform}\left(\{-1,+1\}\right)$, scales $h_q$ by $\mu^s$, and clips the result to $\left[\mathcal{H}_q^{\min},\mathcal{H}_q^{\max}\right]$.
For numeric hyperparameters with $\mathcal{H}_q\subseteq[0,1]$, \textsc{Mutate} instead scales the distance to the nearer boundary, setting $h_q\gets h_q\cdot\mu^s$ if $h_q\le 0.5$ and $h_q\gets 1-\left(1-h_q\right)\cdot\mu^s$ otherwise, before clipping to $\left[\mathcal{H}_q^{\min},\mathcal{H}_q^{\max}\right]$.
If clipping leaves $h_q$ unchanged because it already sat at a bound, \textsc{Mutate} applies the opposite sign $-s$ to the original value and clips again.

For integer hyperparameters, \textsc{Mutate} rounds the scaled value and then rounds up to the parameter's required multiple before clipping.
The multiple is the number of learning GPUs ($4$) for scenarios per iteration and minibatch size, $10$ for the Gaia2 and AppWorld generation limits, and $1$ otherwise.
If rounding returns the original value, \textsc{Mutate} adds or subtracts one step of the multiple in direction $s$ before clipping, ensuring that a selected integer always changes.

An unset per-family sampling weight defaults to $1.0$; when first selected for mutation, \textsc{Mutate} initializes it at $1.0$ and perturbs it in the same draw.
Finally, if a mutation would invert the GSPO clipping bounds ($\epsilon_{\mathrm{low}}>\epsilon_{\mathrm{high}}$), \textsc{Mutate} sets $\epsilon_{\mathrm{high}}\gets\epsilon_{\mathrm{low}}$.

\clearpage
\subsection{Hyperparameter Visualization}
\label{app:scaling_behavior_vis}

The hyperparameter maps in \cref{fig:behavior_hparams,fig:scaling_16p_hparams} use manually constructed axes inspired by low-dimensional embedding methods such as t-SNE while preserving interpretability.
We project each configuration explored by PBPT onto two directions in hyperparameter space defined by the weights in \cref{tab:hparam-axes}.
Specifically, we apply a $\log_{10}$ transform to each strictly positive numeric hyperparameter, standardize each hyperparameter across all configurations in the run, and compute each axis as the weighted mean of the standardized values normalized by the sum of absolute weights.
Positive weights move a configuration toward either larger optimization steps or more precise, computationally expensive gradient estimates.
We omit weight decay and evaluation settings (zero weight), as they do not directly affect either axis.

\begin{table}[ht]
  \centering
  \caption{Axis weights for the hyperparameter maps in \cref{fig:behavior_hparams,fig:scaling_16p_hparams}.}
  \label{tab:hparam-axes}
  \small
  \setlength{\tabcolsep}{6pt}
\begin{tabular}{@{}lcc@{}}
  \toprule
  Hyperparameter & Step size & Gradient \\
  \midrule
  Learning rate & $+1.00$ & --- \\
  Epochs per iteration & $+0.60$ & --- \\
  Rollouts per scenario & --- & $+0.80$ \\
  Scenarios per iteration & --- & $+1.00$ \\
  Parallel scenario runners & $+0.15$ & --- \\
  Minibatch size & $-0.60$ & $+0.60$ \\
  Max gradient norm & $+0.70$ & --- \\
  Lower clipping parameter $\epsilon_{\mathrm{low}}$ & $+0.35$ & --- \\
  Upper clipping parameter $\epsilon_{\mathrm{high}}$ & $+0.35$ & --- \\
  Truncated importance sampling $c$ & $+0.30$ & $-0.40$ \\
  Clip log-prob grads at $-\epsilon$ & $-0.30$ & --- \\
  REPO-R rescaling & $-0.15$ & --- \\
  REPO-R $\beta_{\max}$ & $-0.15$ & --- \\
  Non-zero-advantage fraction & --- & $+0.50$ \\
  Rollouts fraction & --- & $+0.25$ \\
  Rollouts-per-scenario fraction & --- & $+0.25$ \\
  \bottomrule
\end{tabular}
\end{table}

\clearpage
\section{Additional Results}
\label{app:additional_results}

\subsection{Scaling the Population}
\label{app:scaling_behavior}

This section studies how PBPT behaves with a larger population of $16$ members and analyzes how the population's hyperparameters and training behavior evolve over the course of a run.
\Cref{tab:scaling_16p} compares $16$ members with the $8$-member run from \cref{tab:enigmata_summary}.
The two runs differ only in population size.
The larger population reaches higher accuracy and explores roughly $1.8\times$ as many configurations.

\begin{table}[h]
  \centering
  \setlength{\belowcaptionskip}{4pt}
  \caption{PBPT with $8$ and $16$ members on Enigmata.
  Best val.\ and test are the highest single evaluations ($\%$) of any member within $66$ hours.
  Counts cover the full runs, with events summed over members.
  We deduplicate configurations across members, excluding task-mixture weights and evaluation-only settings.
  Selection events correspond to \act{replace} actions.}
  \label{tab:scaling_16p}
  \setlength{\tabcolsep}{4pt}
  \begin{tabular}{@{}lccccccc@{}}
    \toprule
    Members & \shortstack{Best\\val.} & $\Delta$ & \shortstack{Best\\test} & $\Delta$ & \shortstack{Distinct\\configs} & \shortstack{Mutation\\events} & \shortstack{Selection\\events} \\
    \midrule
    $8$ & $80.35$ & -- & $79.36$ & -- & $205$ & $204$ & $82$ \\
    $16$ & $\mathbf{82.10}$ & $+1.75$ & $\mathbf{80.96}$ & $+1.60$ & $363$ & $388$ & $180$ \\
    \bottomrule
  \end{tabular}
\end{table}

\begin{figure}[h]
  \centering
  \includegraphics[width=0.75\textwidth]{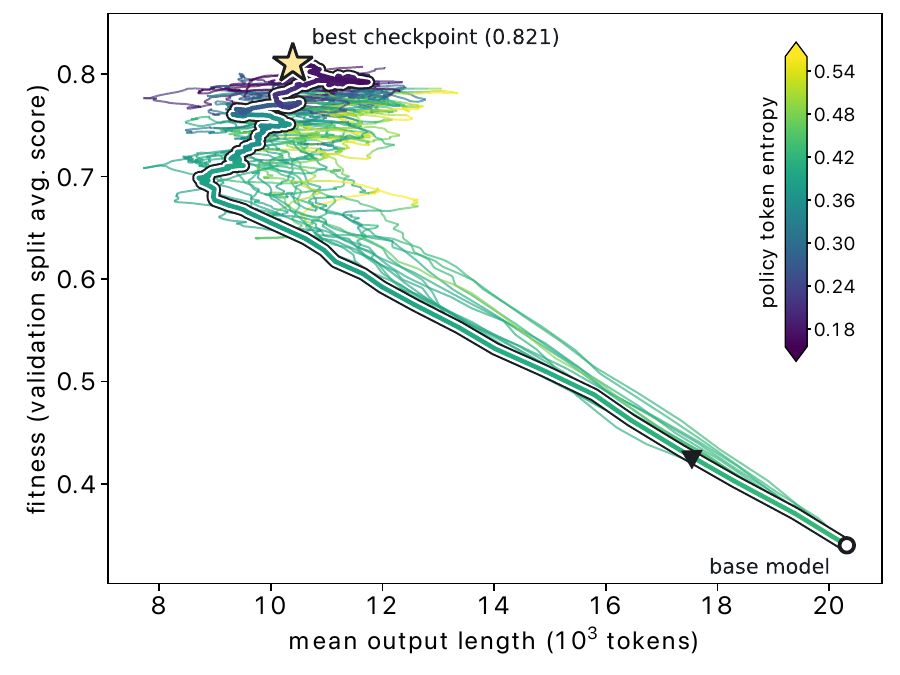}
  \caption{Trajectories of the $16$-member population on Enigmata, colored by policy token entropy.
  The outlined line is the ancestry of the best checkpoint, followed through selection events.}
  \label{fig:scaling_16p_behavior}
\end{figure}

\Cref{fig:scaling_16p_behavior} shows the population's training behavior.
Output length drops quickly from the base model and plateaus, after which accuracy gains track falling policy entropy (Spearman $\rho=-0.71$).

\begin{figure}[ht]
  \centering
  \includegraphics[width=0.60\textwidth]{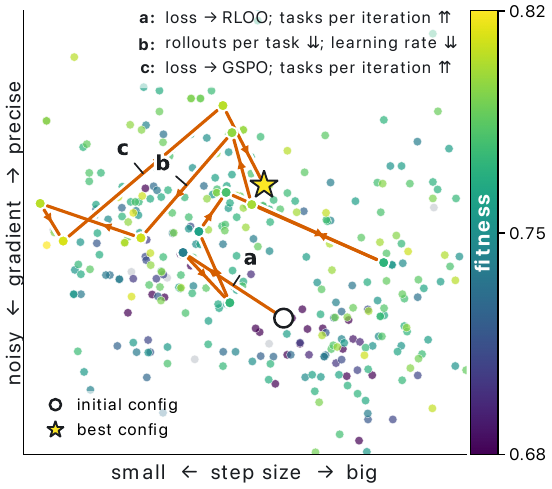}
  \caption{Configurations explored by the $16$-member population, colored by best fitness.
  The line traces the ancestry of the best configuration, with letters marking its notable mutations.}
  \label{fig:scaling_16p_hparams}
\end{figure}

\clearpage
\Cref{fig:scaling_16p_hparams} shows the configurations explored by the $16$-member population, analogously to \cref{fig:behavior_hparams}, using the axes in \cref{app:scaling_behavior_vis}.

\clearpage
\subsection{Nemotron with Qwen3.5-9B}
\label{app:nemotron_9b}

\paragraph{Setup.}
We repeat the Nemotron comparison with Qwen3.5-9B to evaluate PBPT with a larger model on the same benchmark.
Only the base model changes relative to the $4$B comparison.
\Cref{app:nemotron} describes the data splits, verifiers, rewards, and evaluation protocol, and \cref{tab:reasoning-ranges} gives the initial RL configuration and PBPT search space.
Both methods use $8$ learners, each running on $4$ B200 GPUs.
PBPT uses the same initial mutation and mutation settings as the main experiments.
We evaluate the base model separately without an adapter.

\begin{figure}[ht]
  \centering
  \includegraphics[width=\textwidth]{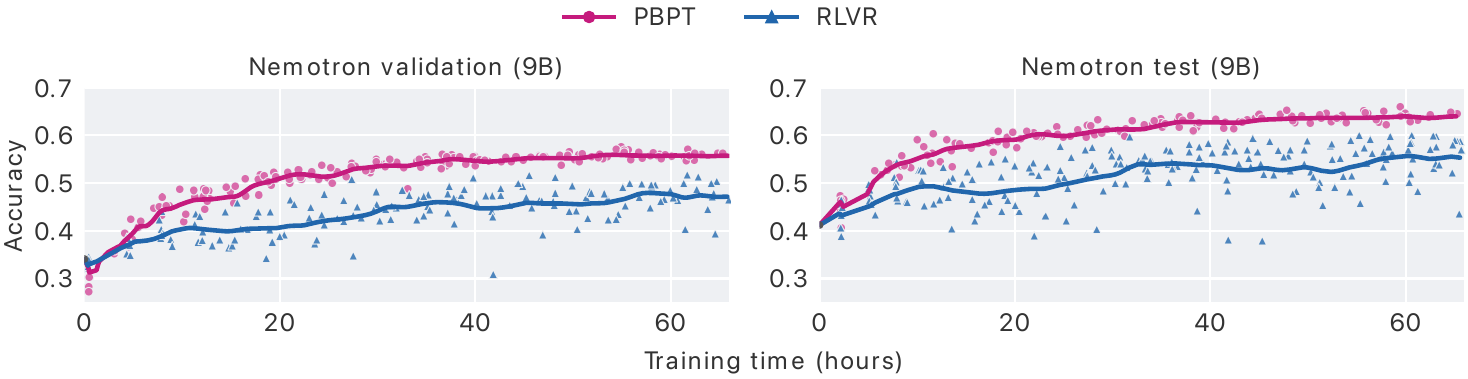}
  \caption{Comparison of PBPT and independent RLVR on Nemotron with Qwen3.5-9B.
  The x-axis shows accumulated training time per learner, and the y-axis shows accuracy.
  Scatter points show subsampled evaluation results, and solid lines show smoothed population means, following the procedure in \cref{app:benchmark_details}.
  The shared point at zero training time reports a separate evaluation of the base model.
  PBPT achieves higher mean validation and test accuracy than independent RLVR.}
  \label{fig:nemotron_9b}
\end{figure}

\paragraph{Results.}
As shown in \cref{fig:nemotron_9b}, PBPT maintains higher mean validation and test accuracy through the later stages of training.
\Cref{tab:nemotron_9b} summarizes test performance using the final-$20\%$ averaging window of \cref{tab:enigmata_summary}.
We first average each learner's evaluations within this window and then give all $8$ learners equal weight.
PBPT produces a stronger population with less variation across members.
Its advantage also holds for learners with the highest and lowest validation means.

\begin{table}[ht]
  \centering
  \caption{Nemotron test accuracy ($\%$) with Qwen3.5-9B, averaged over evaluations in the final $20\%$ of the training budget.
  Avg.\ test reports the mean and standard deviation across the $8$ learners' test means.
  Best-val.\ test and Worst-val.\ test report the test means of the learners with the highest and lowest validation means, respectively.
  Bold marks the highest accuracy in each column.}
  \label{tab:nemotron_9b}
  \begin{tabular}{@{}lccc@{}}
    \toprule
    Method & Avg.\ test & Best-val.\ test & Worst-val.\ test \\
    \midrule
    RLVR & $55.0\pm3.4$ & $59.4$ & $48.6$ \\
    PBPT & $\mathbf{63.8\pm0.6}$ & $\mathbf{64.1}$ & $\mathbf{64.0}$ \\
    \bottomrule
  \end{tabular}
\end{table}

\clearpage
\subsection{Additional Performance Comparisons}
\label{app:additional_performance}

We conduct additional experiments to distinguish the effects of initial configuration diversity, population decisions during training, and inference settings, and to assess consistency across independent runs.
All comparisons use Qwen3.5-4B with $8$ learners per method, each allocated $4$ B200 GPUs and a $66$-hour training budget.
The data splits, rewards, and human-tuned initial RL configurations follow \cref{app:benchmark_details,app:initial_configs}.
The original PBPT and RLVR results are those reported in \cref{subsec:exp_main}.
We use the protocol of \cref{tab:enigmata_summary}, averaging each learner's actual validation and test evaluations over the final $20\%$ of the budget, from $52.8$ to $66$ hours.
Avg.\ test gives equal weight to the $8$ learners' test means, while Best-val.\ test and Worst-val.\ test report the test means of the learners with the highest and lowest validation means.
AppWorld's normal and challenge test sets are reported separately, using the same validation ranking.

\paragraph{Initial configuration diversity.}
We compare full PBPT with an initial-mutation-only variant on Enigmata, Nemotron, and AppWorld.
The variant starts from exactly the same configuration set as full PBPT, including the initial mutations applied to $5$ of the $8$ learners.
Each learner then trains independently without further mutation or training-state transfer.
This also provides a random-search baseline that samples configurations around the human-tuned starting point and trains each for the full budget.
As shown in \cref{tab:initial_mutation_comparison}, full PBPT outperforms this variant across all three benchmarks and all three reported measures.
These results show the benefit of population decisions during training.

\begin{table}[htbp]
  \centering
  \caption{Test performance ($\%$) of independent RLVR, initial-mutation-only PBPT, and full PBPT.
  The two PBPT variants start from identical configuration sets, while RLVR starts from the human-tuned configuration without initial mutation.
  All statistics use the final-$20\%$ averaging protocol of \cref{tab:enigmata_summary}.
  Bold marks the highest value in each column within each benchmark.}
  \label{tab:initial_mutation_comparison}
  \small
  \setlength{\tabcolsep}{5pt}
  \renewcommand{\arraystretch}{1.1}
  \begin{tabular*}{\textwidth}{@{\extracolsep{\fill}}llrrr@{}}
    \toprule
    Benchmark & Method & Avg.\ test & Best-val.\ test & Worst-val.\ test \\
    \midrule
    Enigmata & RLVR & $75.40$ & $78.00$ & $73.03$ \\
     & PBPT (init.\ mut.\ only) & $75.82$ & $77.43$ & $70.86$ \\
     & PBPT (full) & $\mathbf{77.59}$ & $\mathbf{78.12}$ & $\mathbf{77.03}$ \\
    \midrule
    Nemotron & RLVR & $49.28$ & $56.81$ & $44.48$ \\
     & PBPT (init.\ mut.\ only) & $45.89$ & $60.01$ & $36.69$ \\
     & PBPT (full) & $\mathbf{60.57}$ & $\mathbf{62.03}$ & $\mathbf{60.85}$ \\
    \midrule
    AppWorld (normal) & RLVR & $76.99$ & $78.47$ & $65.08$ \\
     & PBPT (init.\ mut.\ only) & $75.61$ & $72.77$ & $66.67$ \\
     & PBPT (full) & $\mathbf{84.73}$ & $\mathbf{86.90}$ & $\mathbf{80.44}$ \\
    \midrule
    AppWorld (challenge) & RLVR & $56.83$ & $\mathbf{68.35}$ & $42.09$ \\
     & PBPT (init.\ mut.\ only) & $52.69$ & $51.14$ & $47.91$ \\
     & PBPT (full) & $\mathbf{63.18}$ & $64.24$ & $\mathbf{67.15}$ \\
    \bottomrule
  \end{tabular*}
\end{table}

\clearpage
\paragraph{Online mutation without state transfer.}
On Nemotron, we evaluate an online-mutation-only variant in which all learners start from the unmutated human-tuned configuration in \cref{tab:reasoning-ranges}.
It uses PBPT's fitness-based mutation rule from \cref{app:implementation_details}, but learners never inherit another member's weights, optimizer state, or hyperparameters.
\Cref{tab:nemotron_additional_comparisons} shows that this variant falls behind full PBPT in all three measures.
Together with the state-transfer-only ablation in \cref{subsec:res_ablation}, these results support combining mutation and state transfer.

\begin{table}[htbp]
  \centering
  \caption{Additional Nemotron comparisons with Qwen3.5-4B.
  Test accuracy ($\%$) follows the final-$20\%$ averaging protocol of \cref{tab:enigmata_summary}.
  The independent rerun changes the random seed, and fixed inference disables temperature mutation.
  Bold marks the highest value in each column.}
  \label{tab:nemotron_additional_comparisons}
  \setlength{\tabcolsep}{7pt}
  \begin{tabular}{@{}lrrr@{}}
    \toprule
    Method & Avg.\ test & Best-val.\ test & Worst-val.\ test \\
    \midrule
    RLVR & $49.28$ & $56.81$ & $44.48$ \\
    PBPT (original) & $\mathbf{60.57}$ & $\mathbf{62.03}$ & $\mathbf{60.85}$ \\
    PBPT (online mut.\ only) & $52.19$ & $57.00$ & $38.57$ \\
    PBPT (rerun) & $59.76$ & $60.06$ & $59.35$ \\
    PBPT (fixed inference) & $58.60$ & $59.41$ & $58.15$ \\
    \bottomrule
  \end{tabular}
\end{table}

\paragraph{Independent repetition.}
We repeat the complete Nemotron PBPT experiment in \cref{subsec:exp_main} with a different random seed.
The rerun uses the same initial RL configuration, initial mutation procedure, and subsequent population decision rules.
The two runs achieve similar performance in \cref{tab:nemotron_additional_comparisons}, providing evidence of consistency across independent populations.

\paragraph{Fixed inference temperature.}
We rerun PBPT on Nemotron with the evaluation temperature fixed at $0.6$, keeping the remaining search space unchanged.
The temperature is excluded from both initial and online mutation, while selection and mutation of the remaining hyperparameters follow \cref{app:implementation_details,tab:reasoning-ranges}.
As in all comparisons, PBPT and independent RLVR have the same limit on the total number of tokens generated per task.
\Cref{tab:nemotron_additional_comparisons} shows that PBPT outperforms independent RLVR with a fixed evaluation temperature.

\clearpage
\paragraph{AppWorld ablations.}
We also evaluate the online-mutation-only and fixed-inference variants on AppWorld, using the same resources and final-$20\%$ averaging protocol as the comparisons above.
The online-mutation-only variant starts from the unmutated human-tuned configuration in \cref{tab:agentic-ranges} and disables training-state transfer, retaining only PBPT's fitness-based mutations for underperforming learners.
As shown in \cref{tab:appworld_additional_comparisons}, full PBPT outperforms this variant across all three measures on both test splits, consistent with the Nemotron results.

For the fixed-inference variant, we fix the evaluation temperature at $0.1$ and the per-turn generation limit at $8{,}192$ tokens, keeping the remaining search space unchanged.
The total evaluation token budget per task remains the same as in full PBPT.
This variant achieves slightly lower mean test performance than full PBPT on both splits, as also observed on Nemotron.

\begin{table}[htbp]
  \centering
  \caption{Additional AppWorld comparisons with Qwen3.5-4B.
  Test performance ($\%$) follows the final-$20\%$ averaging protocol of \cref{tab:enigmata_summary}, using the same validation ranking for both test splits.
  Bold marks the highest value in each column within each split.}
  \label{tab:appworld_additional_comparisons}
  \small
  \setlength{\tabcolsep}{6pt}
  \begin{tabular}{@{}llrrr@{}}
    \toprule
    Split & Method & Avg.\ test & Best-val.\ test & Worst-val.\ test \\
    \midrule
    Test-normal & PBPT (full) & $\mathbf{84.73}$ & $\mathbf{86.90}$ & $80.44$ \\
     & PBPT (online mut.\ only) & $77.77$ & $81.55$ & $77.81$ \\
     & PBPT (fixed inference) & $82.77$ & $84.05$ & $\mathbf{81.02}$ \\
    \midrule
    Test-challenge & PBPT (full) & $\mathbf{63.18}$ & $\mathbf{64.24}$ & $\mathbf{67.15}$ \\
     & PBPT (online mut.\ only) & $56.53$ & $54.00$ & $64.41$ \\
     & PBPT (fixed inference) & $60.33$ & $62.64$ & $57.47$ \\
    \bottomrule
  \end{tabular}
\end{table}

\end{document}